\documentclass{article}
\usepackage{arxiv}

\usepackage{geometry}
\usepackage{xcolor}
\usepackage[hidelinks]{hyperref}
\usepackage{graphicx}
\usepackage{amsmath}
\usepackage{amssymb}
\usepackage{bm}
\usepackage{array}
\usepackage{caption}
\usepackage{url}
\usepackage{subcaption}
\usepackage{algorithm, algpseudocode}
\makeatletter
\let\ftype@table\ftype@figure
\makeatother
\usepackage{cite}
\title{Variational Bayes Deep Operator Network: A data-driven Bayesian solver for parametric differential equations}
\author{
  Shailesh Garg  \\
  Department of Applied Mechanics\\
  Indian Institute of Technology Delhi\\
  Hauz Khas, New Delhi 110016, India. \\
  \texttt{shaileshgarg96@gmail.com} \\
  \And
  Souvik Chakraborty  \\
  Department of Applied Mechanics\\
  Yardi School of Artificial Intelligence (YScAI)\\
  Indian Institute of Technology Delhi\\
  Hauz Khas, New Delhi 110016, India. \\
  \texttt{souvik@am.iitd.ac.in}}
\begin{document}
\maketitle
\begin{abstract}
Neural network based data-driven operator learning schemes have shown tremendous potential in computational mechanics. 
DeepONet is one such neural network architecture which has gained widespread appreciation owing to its excellent prediction capabilities.
Having said that, being set in a deterministic framework exposes DeepONet architecture to the risk of overfitting, poor generalization and in its unaltered form, it is incapable of quantifying the uncertainties associated with its predictions.
We propose in this paper, a Variational Bayes DeepONet (VB-DeepONet) for operator learning, which can alleviate these limitations of DeepONet architecture to a great extent and give user additional information regarding the associated uncertainty at the prediction stage.
The key idea behind neural networks set in Bayesian framework is that, the weights and bias of the neural network are treated as probability distributions instead of point estimates and, Bayesian inference is used to update their prior distribution.
Now, to manage the computational cost associated with approximating the posterior distribution, the proposed VB-DeepONet uses \textit{variational inference}.
Unlike Markov Chain Monte Carlo schemes, variational inference has the capacity to take into account high dimensional posterior distributions while keeping the associated computational cost low.
Different examples covering mechanics problems like diffusion reaction, gravity pendulum, advection diffusion have been shown to illustrate the performance of the proposed VB-DeepONet and comparisons have also been drawn against DeepONet set in deterministic framework.
\end{abstract}
\section{Introduction}
Differential equations \cite{braun1983differential,zwillinger1998handbook} (DE) are among the fundamental mathematical tools of mechanics, which are used extensively by engineers and scientists to model and predict the physical space around us.
From modelling natural phenomenons like weather \cite{bjerknes1999problem,iversen2016short} and fluid flow in streams \cite{batchelor2000introduction,white2006viscous} to modelling artificial constructs like buildings \cite{paz2012structural}, DE are used for a wide array of applied mechanics and engineering problems.
Now, understanding the processes is one thing, but to leverage our knowledge of the same, we need to be able to analyze and solve the underlying DEs.
To this end, several powerful integration schemes \cite{butcher1976implementation,kloeden1992higher} and finite modeling schemes \cite{christie1976finite,hawken1991review,peiro2005finite} have been developed in the past.
These techniques although can converge to the ground truth arbitrarily well, the associated computational cost can be huge especially for complex physical processes, leading to very little practical use; especially for applications where the same system is to be analyzed under varied external influences, for example, design optimization \cite{chen1997reliability,mishra2013reliability,dubourg2011reliability}, reliability analysis \cite{rackwitz2001reliability,alibrandi2015new,sudret2012meta,sudret2017surrogate,bhattacharyya2021structural} and active control problems \cite{datta2003state,fan2020reinforcement}.

To overcome this hurdle, researches have taken advantage of Neural Networks \cite{raissi2019physics,pang2019fpinns,kharazmi2019variational} (NN), which are a subset of machine learning \cite{jordan2015machine,wang2016machine} (ML) algorithms.
NNs have found their use-case in fields of pattern recognition \cite{basu2010use}, speech recognition \cite{deng2013new} and other various industries \cite{abiodun2018state}.
One of the main advantage of using NN algorithms is that, once trained, they can generalize for new environments with little computational cost.
For the purposes of learning the underlying DE/operator, NN algorithms used can be rooted in physics like Physics Informed Neural Networks \cite{raissi2019physics,pang2019fpinns,kharazmi2019variational} or they can be data driven \cite{tripura2022wavelet,zhumekenov2019fourier,lu2021learning,rasmussen2003gaussian,mackay1998introduction}.
Physics based techniques while robust lack the capacity to train for dynamical systems with ambiguity/uncertainty \cite{der2009aleatory,mezic2008uncertainty,bae2004epistemic,garg2022physics,pawar2021nonintrusive,bhattacharyya2019uncertainty} in the known governing DEs.
Data driven techniques on the other hand aim to learn the underlying operator based on the data alone and hence are more flexibly compared to physics based approaches.
This dependency on data alone and being set in a deterministic framework, leaves the data driven techniques vulnerable to issues like overfitting and poor generalization.
To overcome these issues, Bayesian inference \cite{dempster1968generalization,blundell2015weight,box2011bayesian} can be used within the NN framework making the whole architecture robust.
Working in Bayesian framework also allows the user to quantify prediction uncertainties, coming due to the limited nature of training data.

Our focus of study in this paper is on DeepONet \cite{lu2021learning}, which is a deep NN architecture and has gained popularity \cite{wang2021learning,goswami2022physics,di2021deeponet,garg2022assessment} owing to its versatility and performance across varied operator learning problems.
DeepONet is based on principles of the universal approximation theorem \cite{chen1995universal} and it does function to function mapping between the input and output of DE under consideration.
It has one branch and one trunk network which take input function and output location as their inputs respectively and are then combined to obtain the target output.
DeepONet however, like other data driven techniques, lack the capacity to reproduce the ground truth in its entirety and, thus knowing the uncertainty associated with its predictions can give the users an additional confidence in its applications.
Hence, it is worthwhile to explore a Bayesian framework for the Deterministic DeepONet (D-DeepONet), as it not only enables the capacity to quantify uncertainty but also reduces the risk of overfitting at the training stage.

Bayesian Neural Networks \cite{bishop1997bayesian,jospin2022hands,kononenko1989bayesian} (BNN) are based on the principles of Bayes theorem \cite{wasserman2004bayesian} and the key difference in BNNs compared to traditional NNs is that the weights and biases of an BNN are probability distributions instead of point estimates like in a traditional NN.
Because BNN algorithms integrate out the network parameters, that is why they are \textit{not} prone to over-fitting unlike a traditional NN, which is highly susceptible to over-fitting.
Now, in order to train a BNN, among various techniques, two standout approaches are Markov Chain Monte Carlo \cite{brooks1998markov,papamarkou2019challenges} (MCMC) based schemes and Variational Inference \cite{blei2017variational,kingma2013auto,graves2011practical,blundell2015weight} (VI) based schemes.
MCMC although accurate, is computationally expensive and is impractical in the face of large number of training parameters; VI however is an approximate solution and is capable of producing reasonable good results for NNs with large number of trainable parameters.
In the existing literature, DeepONet architectures \cite{lin2021accelerated, moya2022deeponet} based in Bayesian framework exists, but they utilize MCMC based schemes to obtain the posterior distribution which as discussed is intractable in the face of large number of parameters.
In \cite{moya2022deeponet,yang2022scalable}, authors have developed probabilistic frameworks for quantifying uncertainty in DeepONet prediction.
Their methodology however fails to capture fully, the other associated benefits which comes with working in a Bayesian framework.
In this work, we propose for the first time a Variational Bayes based DeepONet (VB-DeepONet) framework that is Bayesian and uses variational inference for estimating the probability density function of the neural network parameters.
The salient features of the proposed approach includes,
\begin{itemize}
    \item \textbf{{Accuracy and better generalization}:} Compared to a frequentist framework, a Bayesian approach is less prone to over-fitting. This is because the unknown weights and biases are integrated out during the prediction stage. Therefore, the proposed VB-DeepONet, as compared to its deterministic counterpart, is expected to yield more accurate prediction and generalize better to unseen data. This is particularly important as the overarching goal in operator learning is to use to trained model for making prediction. 
    \item \textbf{{Uncertainty quantification}:} In data-driven approach, one of the key challenges is to quantify the uncertainty in the learned model because of limited and noisy data. VB-DeepONet being a Bayesian framework has the inherent capability to quantify epistemic uncertainty because of limited and noisy data. This is especially important and helps a designer in making an informed decision.
    \item{{\textbf{Efficiency}}:} The proposed VB-DeepONet is based on variational Bayes and hence, is computationally efficient as compared to a Monte Carlo based approach. In essence, a VB-DeepONet provides a balance between a relatively efficient D-DeepONet (with no uncertainty measure) and an computational expensive Monte Carlo based Bayesian DeepONet.
\end{itemize}
Different examples have been carried out to test the proposed Variational Bayes DeepONet (VB-DeepONet) and the result produced showcase the excellent predictive capabilities of the framework, along with its capacity to quantify the associated uncertainty.

The rest of the paper is arranged as follows. Section \ref{section: deeponet background}
discusses the existing D-DeepONet architecture and Section \ref{section: vbd} details the proposed framework for the VB-DeepONet.
Section \ref{section: numerical illustrations} discusses the examples carried out to test the proposed VB-DeepONet and section \ref{section: conclusion} concludes the findings in this paper.
\section{DeepONet Background}\label{section: deeponet background}
DeepONet is a neural network architecture designed for learning nonlinear operator $G(\bm u)(\bm y)$ based on the training data $\mathbf D = \{\bm u^{(i)}, \bm y^{(i)}, s^{(i)}\}_{i=1}^{N_s}$.
To achieve this, it learns the mapping between the input force $\bm u$ and output $s$ corresponding to location vector $\bm y$.
It is designed based on the universal approximation theorem \cite{chen1995universal} for operator learning, which states that a NN can estimate any nonlinear operator, so long as there is no constraint placed on the depth and width of the NN.
Accordingly, DeepONet's architecture is bifurcated into two parts, (i) Branch net and (ii) Trunk net.
The branch net takes $\bm u$ as input while the trunk net takes $\bm y$ as input. 
Individual nets may be stacked or unstacked neural networks, so long as the outputs from the branch and trunk nets are of same dimension.
The two outputs are then combined using dot product and the resulting output is obtained from the final output node.
Mathematically, this can be described as follows:
\begin{equation}
    \begin{gathered}
    G(\bm u)(\bm y) = \left<O^{(b)}\,,O^{(t)}\right>\\
    O^{(b)} = \alpha^{(b)}\left(b^{(b)}+\sum\limits_ix^{(b)}_iw^{(b)}_i\right)\\
    O^{(t)} = \alpha^{(t)}\left(b^{(t)}+\sum\limits_ix^{(t)}_iw^{(t)}_i\right)
    \end{gathered}
	\label{equation: dot product}
\end{equation}
where $G(\bm u)(\bm y)$ is the target nonlinear operator. Letter $(b)$ represents the branch network and $(t)$ represents the trunk network. $O^{(\cdot)}$ is the output of the last layer, $\alpha^{(\cdot)}$ is the activation function, and $b^{(\cdot)}$ is the bias in the last layer.
$x_i^{(\cdot)}$ are the neurons with weights $w_i^{(\cdot)}$ in the last layer.
$\left<\cdot,\cdot\right>$ represents the dot product.
The weights and bias in Eq. \eqref{equation: dot product} are point estimates and therefore the D-DeepONet cannot account for the epistemic uncertainties coming due to limited and noisy data.
\section{Proposed VB-DeepONet}\label{section: vbd}
\begin{figure}[ht!]
    \centering
    \includegraphics[width = 0.85\textwidth]{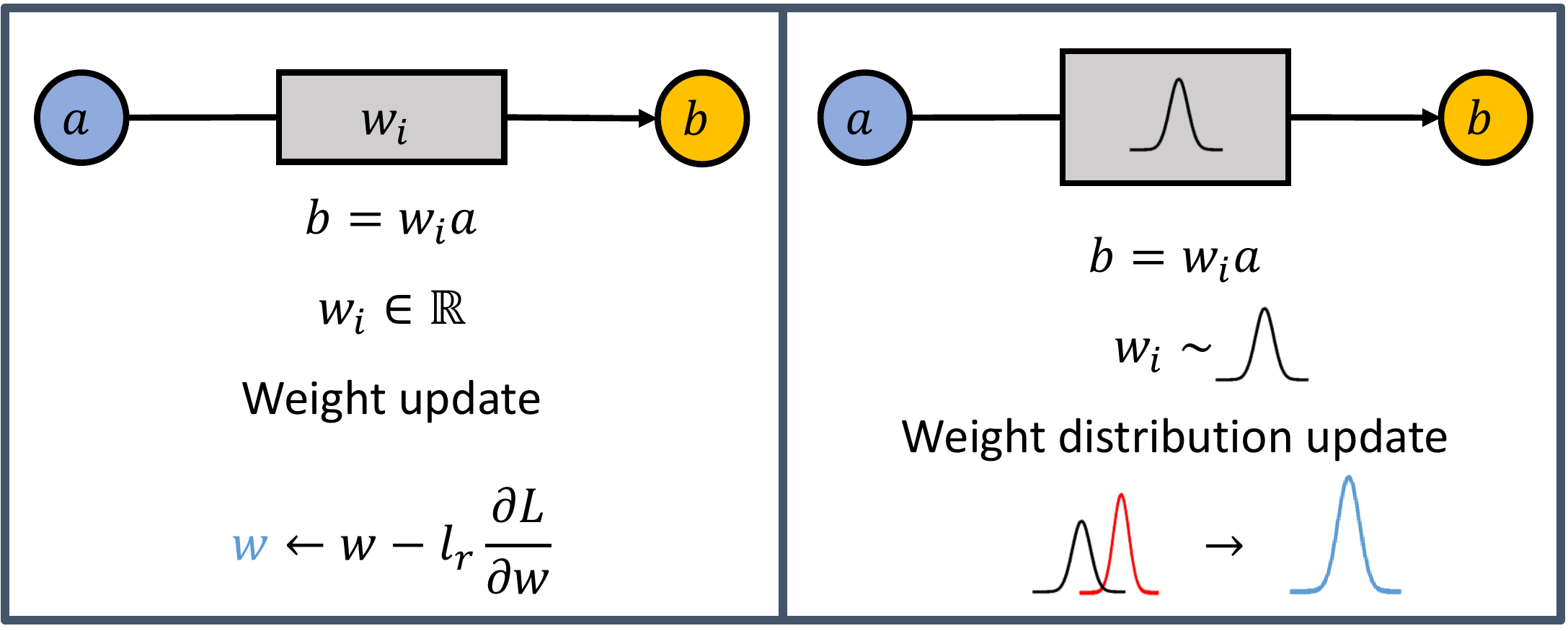}
    \caption{Parameter update in deterministic neural network(left) and in Bayesian neural network (right). 
    The update for prior distribution, shown in red is the normalized likelihood obtained based on the training data.
    }
    \label{fig: BNN schematic}
\end{figure}

Bayesian neural networks remedy the aforementioned limitations of D-DeepONet by considering the network parameters as probability distributions and updating them based on the principles of Bayesian inference.
This enables the network architecture to account for  epistemic uncertainties due to limited and noisy data.
A parallel approach in certain aspects to BNNs in deterministic framework could be ensemble training; but it is computationally very expensive to train multiple copies of same neural network especially for dense architectures like D-DeepONet.
A schematic showing the key difference in BNN and a NN  is shown in Fig. \ref{fig: BNN schematic}
\subsection{Variational Bayes DeepONet}
The D-DeepONet architecture aims to learn the underlying operator based on the training data and the same can be represented as:
\begin{equation}
    s = G(\bm u)(\bm y) = N_D(\bm u,\bm y,\bm\theta_{t}),
\end{equation}
where $N_D(\bm u,\bm y,\bm\theta_{t})$ represents the output of the D-DeepONet architecture and, it is a function of input $\bm u$, location point $\bm y$ and deterministic trainable NN parameters $\bm\theta_{t}$.
The optimized $\bm\theta_{t}$ obtained after training the D-DeepONet are point estimates and thus there is no scope for uncertainty quantification.
One way to remedy this could be to train an ensemble of D-DeepONet with same training data but different initial parameters.
The final prediction ensemble thus obtained could be used to quantify the uncertainty associated with its predictions.
If the variation among different predictions corresponding to same training data is more, uncertainty will be more and if the predictions are closer to each other, uncertainty will be less.
But as discussed earlier, training an ensemble of D-DeepONet architectures carries with it, huge computational cost.
BNNs provides a rigorous framework that uses Bayes rule to compute the posterior distribution of the parameters. 
Going forward, the uncertain trainable NN parameters, sampled from a probability distribution will be denoted as $\bm\theta_{tv}$.
To update the prior knowledge about the parameters $\bm\theta_{tv}$, based on the training data  $\mathbf D = \{\bm u^{(i)}, \bm y^{(i)}, s^{(i)}\}_{i=1}^{N_s}$, Bayes theorem is used and the resulting posterior $p(\bm\theta_{tv}|\mathbf D)$ is evaluated as:
\begin{equation}
    p(\bm\theta_{tv}|\mathbf D) = \dfrac{p(\mathbf D|\bm\theta_{tv})p(\bm\theta_{tv})}{\int p(\mathbf D|\bm\theta_{tv}) p(\bm\theta_{tv})d\bm\theta_{tv}},
    \label{equation: bayesian update}
\end{equation}
where $p(\mathbf D|\bm\theta_{tv})$  is the data-likelihood and $p(\bm\theta_{tv})$ is the prior for the trainable parameters.
The denominator $\int p(\mathbf D|\bm\theta_{tv}) p(\bm\theta_{tv})d\bm\theta_{tv}$ in Eq. \eqref{equation: bayesian update} is a normalizing constant and is oftern intractable.
\subsubsection{Prior: VB-DeepONet Parameters}
Modelling the NN parameters as probability distributions allows user to give a more generalized estimate for the parameter instead of a fixed point estimate as given in D-DeepONet and; in the proposed VB-DeepONet, we use the standard normal distribution as prior for the trainable parameters i.e.
\begin{equation}
    p(\bm\theta_{tv}) = \mathcal N(\bm 0, \mathbb I).
\end{equation}
We note that the prior is generally meant to represent our prior knowledge about the system parameters; however, in the absence of any prior knowledge, selecting standard normal distribution as prior is a common practice in literature \cite{murphy2012machine}. We also note that the proposed algorithm is generic and can be used with other priors as well. 
\subsubsection{Likelihood}
The data-likelihood $p(\mathbf D|\bm \theta_{tv})$ can be computed as:
\begin{equation}
    p(\mathbf D|\bm \theta_{tv}) =
    \prod\limits_{i = 1}^{N_s} p(s^{(i)}|\bm u^{(i)}, \bm y^{(i)}, \bm\theta_{tv}),
    \label{equation: likelihood}
\end{equation}
where $p(s^{(i)}|\bm u^{(i)}, \bm y^{(i)}, \bm\theta_{tv})$ is the distribution for the $i-$th target output (in the training data) given the inputs and NN parameters.
Note that Eq. \eqref{equation: likelihood} is based on the fact that the training samples are assumed to be independent and identically distributed.
Now, the output in the proposed VB-DeepONet is modelled as a normal distribution with mean $\mu$ and standard deviation $\sigma$, which themselves follow a probability distribution. The joint probability of the output $s^{(i)}$, mean $\mu$ and standard deviation $\sigma$ can be discretized as:
\begin{equation}
\begin{aligned}
    p(s^{(i)},\mu^{(i)},\sigma^{(i)}|\bm u^{(i)}, \bm y^{(i)}, \bm\theta_{tv}) & = p(s^{(i)}|\mu^{(i)},\sigma^{(i)})p(\mu^{(i)}, \sigma^{(i)}|\bm u^{(i)}, \bm y^{(i)}, \bm\theta_{tv})\\
    & = \mathcal N(\mu^{(i)},{\sigma^{(i)}}^2)p(\mu^{(i)}, \sigma^{(i)}|\bm u^{(i)}, \bm y^{(i)}, \bm\theta_{tv}).
\end{aligned}
\label{equation: joint probability}
\end{equation}
The second term in Eq. \eqref{equation: joint probability} is not known a-priori, and is instead modelled using DeepONet with random weights and biases.
Therefore one can generate samples for $\mu^{(i)}$ and $\sigma^{(i)}$ (refer Fig. \ref{figure: schematic VB-DeepONet}).
Accordingly, Eq. \eqref{equation: joint probability} is represented as:
\begin{equation}
    p(s^{(i)},\mu^{(i)},\sigma^{(i)}|\bm u^{(i)}, \bm y^{(i)}, \bm\theta_{tv}) = \mathcal N(\mu^{(i)},{\sigma^{(i)}}^2)q_{VB}(\mu^{(i)}, \sigma^{(i)}|\bm u^{(i)}, \bm y^{(i)}, \bm\theta_{tv}),
\label{equation: joint probability modified}
\end{equation}
where $q_{VB}(\mu^{(i)}, \sigma^{(i)}|\bm u^{(i)}, \bm y^{(i)}, \bm\theta_{tv})$ represents the DeepONet parameterized by probabilistic $\bm\theta_{tv}$.
Using Eq. \eqref{equation: joint probability modified}, the marginal distribution $p(s|\bm u, \bm y, \bm\theta_{tv})$ is computed as:
\begin{equation}
    p(s^{(i)}|\bm u^{(i)}, \bm y^{(i)}, \bm\theta_{tv}) = \iint \mathcal N(\mu^{(i)},{\sigma^{(i)}}^2)q_{VB}(\mu^{(i)}, \sigma^{(i)}|\bm u^{(i)}, \bm y^{(i)}, \bm\theta_{tv})\,d\mu^{(i)}\,d\sigma^{(i)}
\label{equation: joint prob}
\end{equation}
Using Eq. \eqref{equation: joint prob}, Eq. \eqref{equation: likelihood} is now expressed as:
\begin{equation}
    p(\mathbf D|\theta_{tv}) = \prod\limits_{k = 1}^{N_s}\iint \mathcal N(\mu^{(k)},{\sigma^{(k)}}^2)q_{VB}(\mu^{(k)}, \sigma^{(k)}|\bm u^{(k)}, \bm y^{(k)}, \bm\theta_{tv})\,d\mu^{(k)}\,d\sigma^{(k)}
\label{equation: likelihood final}
\end{equation}
\subsubsection{Variational Inference}
Having described the prior and likelihood for the proposed VB-DeepONet, using Eq. \eqref{equation: bayesian update}, theoretically, the posterior distribution $p(\bm\theta_{tv}|\mathbf D)$ for the parameters $\bm\theta_{tv}$ can be obtained.
However \textit{practically} computing the posterior has following challenges:
\begin{itemize}
    \item An analytical solution can not be generated for the integral because the distributions for parameters $ \mu^{(k)}$ and $ \sigma^{(k)}, \forall k$ are unknown, which contribute to the likelihood distribution $p(\mathbf D|\theta_{tv})$ (refer Eq. \eqref{equation: likelihood final}). This is because  $\int p(\mathbf D|\bm\theta_{tv})p(\bm\theta_{tv})d\bm\theta_{tv}$ is  intractable.
    \item It is possible to use Monte Carlo based approaches to draw samples from the posterior; however, Monte Carlo based approaches are computationally expensive, specifically for a overparameterized network like the DeepONet. 
\end{itemize}
To get around these challenges, variational inference (VI) is used in the proposed VB-DeepONet.
In VI, the posterior distribution $p(\bm\theta_{tv}|\mathbf D)$ to be estimated is approximated by a variational distribution $q(\bm\theta_{tv}|\bm\theta)$, where $\bm\theta$ represents the parameters of variational distribution.
The parameters of the variational distribution are estimated by minimizing the distance between the posterior and the variational distributions. Training the VB-DeepONet entails minimizing this distance. Although there exists a number of distance metrics in the literature, we have used KL divergence \cite{kullback1951information} in this work.
Accordingly, the loss function for the VB-DeepONet can then be represented as:
\begin{equation}
\begin{aligned}
    L &= KL(q(\bm\theta_{tv}|\bm\theta)\,||\,p(\bm\theta_{tv}|\mathbf D))\\[0.25cm]
    &= \int log\left[\dfrac{q(\bm\theta_{tv}|\bm\theta)}{p(\bm\theta_{tv}|\mathbf D)}\right]q(\bm\theta_{tv}|\bm\theta)\,d\bm\theta_{tv}\\[0.25cm]
    &= \int log\left[\dfrac{q(\bm\theta_{tv}|\bm\theta)}{p(\mathbf D|\bm\theta_{tv})p(\bm\theta_{tv})}p(\mathbf D)\right]q(\bm\theta_{tv}|\bm\theta)\,d\bm\theta_{tv}\\[0.25cm]
    &= \int log\left[q(\bm\theta_{tv}|\bm\theta)-log\,p(\bm\theta_{tv})-log\,p(\mathbf D|\bm\theta_{tv})+log\,p(\mathbf D)\right]q(\bm\theta_{tv}|\bm\theta)\,d\bm\theta_{tv}
\end{aligned}
\end{equation}
In the above equation, the term $log\,p(\mathbf D)$ is independent of variational distribution and NN parameters and thus can be ignored in the final calculation.
The final form of the loss function can then be written as:
\begin{equation}
\begin{aligned}
    L &= \int log\left[q(\bm\theta_{tv}|\bm\theta)-log\,p(\bm\theta_{tv})-log\,p(\mathbf D|\bm\theta_{tv})\right]q(\bm\theta_{tv}|\bm\theta)\,d\bm\theta_{tv}\\
    &=\int log\left[q(\bm\theta_{tv}|\bm\theta)\right]q(\bm\theta_{tv}|\bm\theta)\,d\bm\theta_{tv}-\int log\left[p(\bm\theta_{tv})\right]q(\bm\theta_{tv}|\bm\theta)\,d\bm\theta_{tv}-\int log\left[p(\mathbf D|\bm\theta_{tv})\right]q(\bm\theta_{tv}|\bm\theta)\,d\bm\theta_{tv}
\label{equation: loss function}
\end{aligned}    
\end{equation}
The first two terms in the loss function are dependent on the prior selected for NN parameters and their combined effect is termed as the complexity cost. The third term in the loss function is data dependent, which is termed as likelihood cost \cite{blundell2015weight}.
The likelihood cost in the loss function for the proposed VB-DeepONet can be written as:
\begin{multline}
    \int log\left[p(\mathbf D|\bm\theta_{tv})\right]q(\bm\theta_{tv}|\bm\theta)\,d\bm\theta_{tv} =\\ \int log\,\left[\prod\limits_{k = 1}^{N_s}\iint \mathcal N(\mu^{(k)},{\sigma^{(k)}}^2)q_{VB}(\mu^{(k)}, \sigma^{(k)}|\bm u^{(k)}, \bm y^{(k)}, \bm\theta_{tv})\,d\mu^{(k)}\,d\sigma^{(k)}\right]q(\bm\theta_{tv}|\bm\theta)\,d\bm\theta_{tv}
\label{equation: log likelihood}
\end{multline}
Eq. \eqref{equation: log likelihood} can be re-written as:
\begin{multline}
    \int log\left[p(\mathbf D|\bm\theta_{tv})\right]q(\bm\theta_{tv}|\bm\theta)\,d\bm\theta_{tv} =\\ \dfrac{1}{N_t}\dfrac{1}{N}\dfrac{1}{N}\sum\limits_{l = 1}^{N_t}\sum\limits_{k = 1}^{N_s}\sum\limits_{i = 1}^{N}\sum\limits_{j = 1}^{N}log\,\mathcal N(\mu_i(\bm u^{(k)}, \bm y^{(k)}, \bm\theta_{tv}^{(l)}),\sigma_j(\bm u^{(k)}, \bm y^{(k)}, \bm\theta_{tv}^{(l)}))
\label{equation: log likelihood rewritten}
\end{multline}
Practically, while implementing the loss function, Eq. \eqref{equation: log likelihood rewritten} can be simplified as follows:
\begin{equation}
    E_{q(\bm\theta_{tv}|\bm\theta)}\left[log\,p(\mathbf D|\bm\theta_{tv})\right] = \dfrac{1}{\widetilde N}\sum\limits_{l = 1}^{\widetilde N}\sum\limits_{k = 1}^{N_s}log\,\mathcal N(\mu(\bm u^{(k)}, \bm y^{(k)}, \bm\theta_{tv}^{(l)}),\sigma(\bm u^{(k)}, \bm y^{(k)}, \bm\theta_{tv}^{(l)}))
\end{equation}
The above simplification is justified by the fact that the parameters $\mu$ and $\sigma$ are functions of trainable parameters $\bm\theta_{tv}$ and hence taking summation over several values of $\bm\theta_{tv}$ will automatically integrate out $\mu$ and $\sigma$, as is the target in Eq. \eqref{equation: joint prob}.
In the available literature \cite{jospin2022hands,graves2011practical}, $\widetilde N = 1$ has been shown to produce satisfactory results, but in order to improve stability of the whole framework, we have taken $\widetilde N = 25$.
This while increases the computational cost per epoch, the overall number of epochs required reduces hence balancing the computational cost to some extent. 
In the proposed VB-DeepONet, the variational distribution $q(\bm\theta_{tv}|\bm\theta)$ is modelled as a normal distribution with mean vector $\bm\mu_v$ and standard deviation vector $\bm\sigma_v$ and is represented as:
\begin{equation}
    q(\bm\theta_{tv}|\bm\theta) = \mathcal N(\bm\mu_v,diag(\bm\sigma_v)^2),\,\,\,\,\,\,\,\,\,\,\bm\theta = \{\mu_v,\sigma_v\}
\end{equation}
\subsubsection{Implementation}
\begin{figure}[ht!]
    \centering
    \includegraphics[width = 0.95\textwidth]{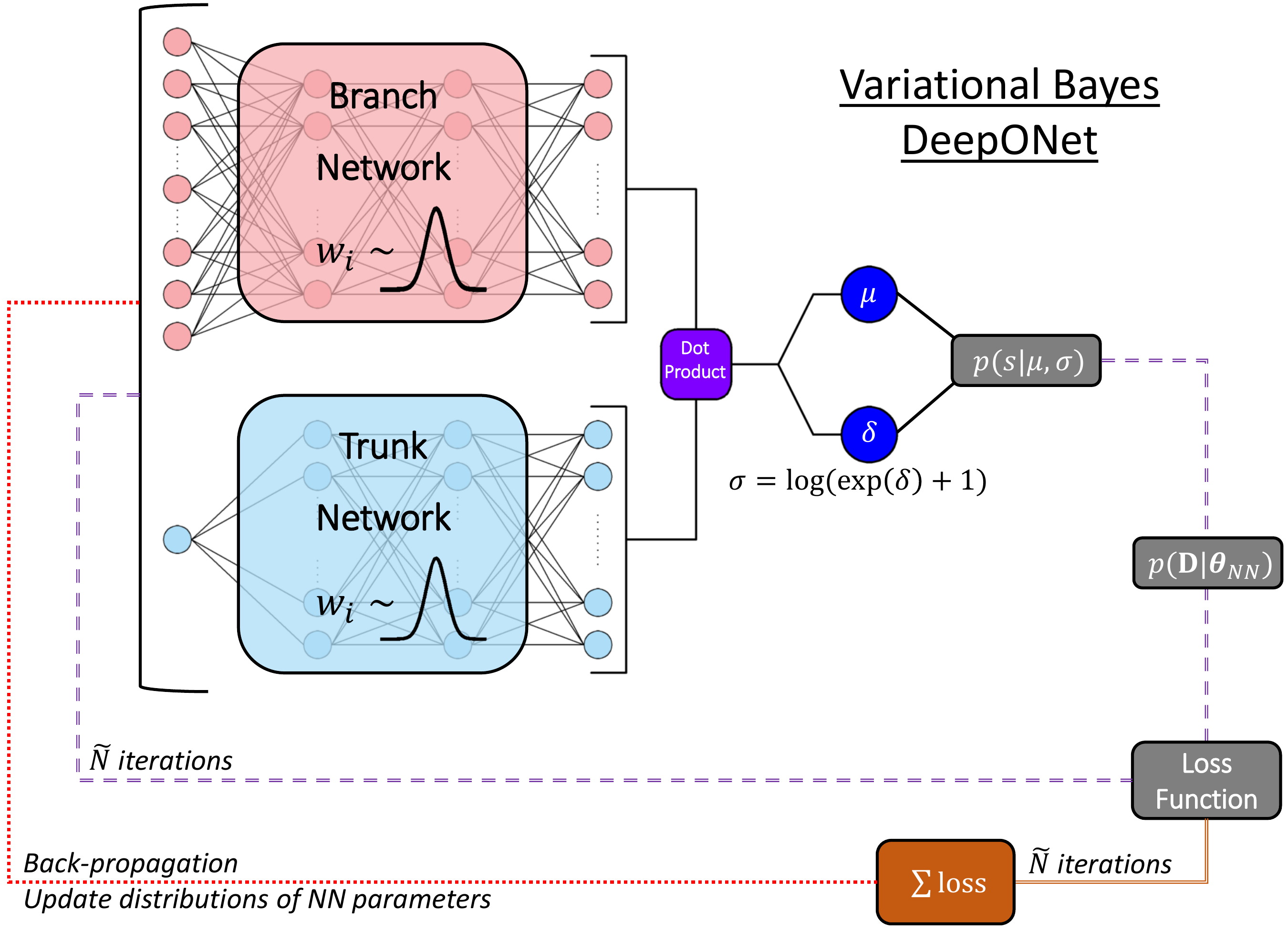}
    \caption{Schematic for proposed VB-DeepONet using variational inference.}
    \label{figure: schematic VB-DeepONet}
\end{figure}

Now that we have a theoretical setup and the loss function for the proposed VB-DeepONet, we can look at the implementation of the VB-DeepONet architecture.
The NN architecture for VB-DeepONet is similar to D-DeepONet with the difference being that in the last step, instead of getting a single value from the dot product of outputs of branch and trunk networks, the dot product is passed onto two densely connected output nodes with uncertain weights and linear activation function (refer Fig. \ref{figure: schematic VB-DeepONet}).

The mean and standard deviation for the distribution $p(s|\mu, \sigma)$ are obtained from the two final output nodes of VB-DeepONet respectively.
For numerical stability, instead of directly getting the standard deviation $\sigma$ from the VB-DeepONet, another parameter $\delta$ is extracted and standard deviation is computed as:
\begin{equation}
    \sigma = log\,(exp\,(\delta) + 1).
\end{equation}
As for branch net and trunk net in VB-DeepONet, there individual architectures will depend on the complexity of operator being learnt. 
One we have finalized the architecture, NN parameters $\bm\theta_{tv}$ will be initialized as defined in the previous subsections, $p(\bm\theta_{tv}) = \mathcal N(\bm 0,\mathbb I)$.
While training the VB-DeepONet model, to make possible the optimization of trainable parameters' distribution, a reparameterization trick \cite{opper2009variational,kingma2013auto,rezende2014stochastic} is employed.
In this, a sample $\kappa$ is drawn from a distribution with known parameters ($\mathcal N(0,1)$), and it is modified using a deterministic function (for which optimization is trivial) to match a sample drawn from the posterior distribution of the training parameter.
The sample for the trainable parameter in the proposed VB-DeepONet is thus represented as:
\begin{equation}
    \theta_{tv}^{(i)} = \mu_v^{(i)}+\sigma_v^{(i)}\kappa, 
    \label{equation: weight sampling}
\end{equation}
where $\theta_{tv}^{(i)}$ is the NN parameter for the $i^{\text{th}}$ node of the VB-DeepONet.
Again, for numerical stability, $\sigma_v^{(i)}$ in Eq. \eqref{equation: weight sampling} is replaced by the expression $log\,(exp\,(\delta^{(\theta)(i)}) + 1)$.
Hence, the parameters $\bm\theta$ to be optimized using back-propagation becomes the mean $\bm\mu_v$ and parameter $\bm\delta^{(\theta)}$, which will in-turn update the variational distribution $q(\bm\theta_{tv}|\bm\theta)$.
The gradients $\Delta\bm\mu_v$ and $\Delta\bm\delta^{(\theta)}$ for updating $\bm\mu_v$ and $\bm\delta^{(\theta)}$ can be computed as follows:
\begin{equation}
\begin{gathered}
    \Delta\bm\mu_v = \dfrac{\partial L}{\partial \bm\theta_{tv}}+\dfrac{\partial L}{\partial\bm\mu_v}\\
    \Delta\bm\delta = \dfrac{\partial L}{\partial \bm\theta_{tv}}\dfrac{\kappa}{exp(-\bm\delta^{(\theta)})+1}+\dfrac{\partial L}{\partial\bm\delta^{(\theta)}}.
    \label{equation: gradients}
\end{gathered}
\end{equation}
Algorithm \ref{algorithm: VB-DeepONet TP} details the training process of the proposed VB-DeepONet.
To implement the VB-DeepONet using variational inference, \textit{Tensorflow Probability} \cite{dillon2017tensorflow,durr2020probabilistic} package has been used.
\begin{algorithm}[ht!]
\caption{Training algorithm for the VB-DeepONet architecture.}
\label{algorithm: VB-DeepONet TP}
\begin{algorithmic}[1]
\State Collect training data and arrange in the format $\mathbf D = \{\bm u^{(i)}, \bm y^{(i)}, s^{(i)}\}_{i=1}^{N_s}$.
\State Set standard normal distribution as the prior distribution for the NN parameters $\bm\theta_{t}$.
\For{i = 1\,:\,epochs}
\For{j = 1\,:\,$\widetilde N$}
\State Generate samples from $\bm\kappa\sim\mathcal N(0,\mathbb I)$ for reparameterization trick.
\State Compute the NN parameters as, $\theta_{tv}^{(k)} = \mu_v^{(k)}+log\,(exp\,(\delta^{(\theta)(k)}) + 1)\kappa^{(k)}$.
\State Input training data to branch and trunk nets.
\State Pass the output from final two output nodes to the output distribution $p(s|\mu, \sigma)$.
\State Compute the loss function as per Eq. \eqref{equation: loss function}.
\State Add the loss function obtained for the $\widetilde N$ iterations.
\EndFor
\State loss $\leftarrow$ loss/$\widetilde N$
\State Compute the gradients $\Delta\mu_v$ and $\Delta\delta$ using Eq. \eqref{equation: gradients}.
\State $\mu_v\leftarrow\mu_v-l_r\Delta\mu$ and $\delta\leftarrow\delta-l_r\Delta\delta$ 
\EndFor
\State Outcome: Trained VB-DeepONet model.
\end{algorithmic}
\end{algorithm}
\subsubsection{Prediction}
Once the VB-DeepONet model is trained, one may compute the predictive distribution using Bayesian inference as:
\begin{equation}
	p(s'|\bm u', \bm y',\mathbf D) = \int p(s'|\bm u', \bm y',\bm\theta_{tv})p(\bm\theta_{tv}|\mathbf D) d\bm\theta_{tv},
\end{equation}
where $\bm u'$ and $\bm y'$ are the testing/prediction inputs and $s'$ is the expected output.
Practically, we sample multiple outputs corresponding to same inputs and compute the predictive mean and predictive variance in order to  quantify the uncertainties and obtain the confidence intervals.
\section{Numerical Illustration}\label{section: numerical illustrations}
This section covers four different examples, covering both ordinary and partial DEs.
The first example follows an elementary Anti-derivative (AD) operator while the second example deals with a Gravity Pendulum (GP) under the influence of external forces.
The third example covers a Diffusion Reaction (DR) system while the fourth example covers an Advection Diffusion (ADVD) equation wherein mapping is carried out between the initial condition and the output of the partial differential equation.

\begin{figure}[ht!]
    \centering
    \includegraphics[width = 0.75\textwidth]{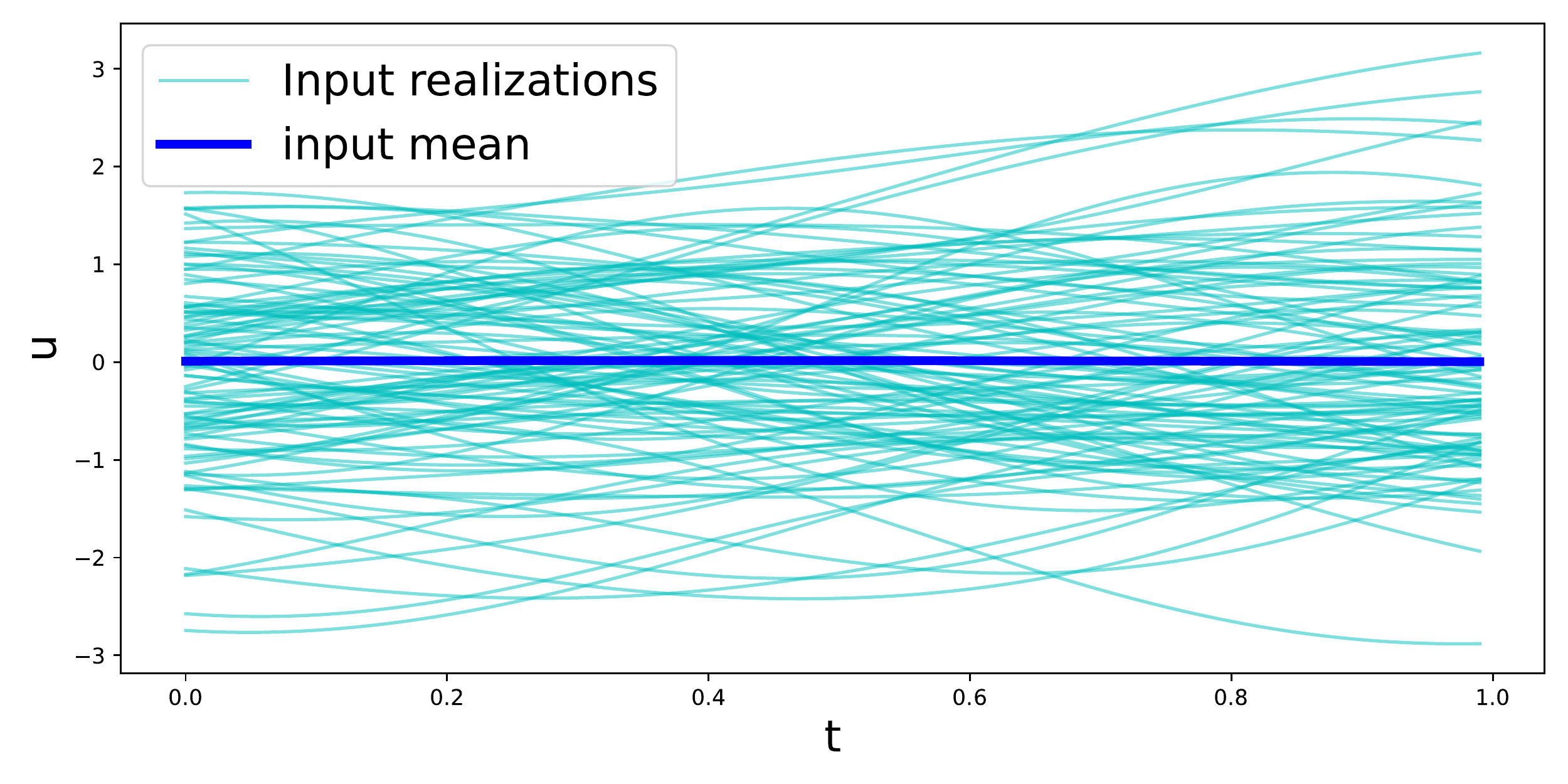}
    \caption{Input Realizations generated using Eq. \eqref{equation: input eqn}.}
    \label{figure: input realizations}
\end{figure}

The inputs $\bm u$ for all the examples is drawn randomly from a Gaussian random field defined as follows:
\begin{equation}
    \begin{gathered}
    u(\bm x)\sim\mathcal{GP}(\bm 0, \kappa(x_i, x_j))\\
    \kappa(x_i,x_j) = exp\left(\dfrac{(x_i-x_j)^2}{2l^2}\right),\,\,\,\,\, x_i,x_j\in\bm x,
    \end{gathered}
    \label{equation: input eqn}
\end{equation}
where $\kappa$ is the covariance matrix and $l = 0.5$ is the length scale for the kernel function.
Fig. \ref{figure: input realizations} shows multiple realizations of the input ensemble along with the mean value.
As can be seen, the realizations have significant variation among them. 
The training dataset for the proposed VB-DeepONet contains multiple samples corresponding to random inputs.
One sample in the training dataset contains $\bm u$ discretized at 100 uniformly spaced points and output $s_i$ corresponding to a random location $\bm y_i$.
$\bm y = \{t\}$ for ordinary DEs in AD and GP examples and, $\bm y = \{x, t\}$ for partial DEs in DR and ADVD examples.

It should also be noted that for one particular $\bm u$, multiple training samples are generated by noting the desired output at random locations $\bm y$.
The input realizations however are discretized at same locations for all the samples.
The training dataset containing samples corresponding to $n$ unique input realizations with $m$ samples corresponding to each unique input can be represented as:
\begin{equation}
    \begin{bmatrix}
    \color{blue} u_1^{(1)} & \color{blue} u_2^{(1)} ..... & \color{blue} u_{99}^{(1)} & \color{blue} u_{100}^{(1)}\\ 
    \color{blue} u_1^{(1)} & \color{blue} u_2^{(1)} ..... & \color{blue} u_{99}^{(1)} & \color{blue} u_{100}^{(1)}\\ 
    \color{blue} .\\
    \color{blue} u_1^{(1)} & \color{blue} u_2^{(1)} ..... & \color{blue} u_{99}^{(1)} & \color{blue} u_{100}^{(1)}\\ 
    \color{red} u_1^{(2)} & \color{red} u_2^{(2)} ..... & \color{red} u_{99}^{(2)} & \color{red} u_{100}^{(2)}\\
    \color{red} u_1^{(2)} & \color{red} u_2^{(2)} ..... & \color{red} u_{99}^{(2)} & \color{red} u_{100}^{(2)}\\
    \color{red} .\\
    \color{red} u_1^{(2)} & \color{red} u_2^{(2)} ..... & \color{red} u_{99}^{(2)} & \color{red} u_{100}^{(2)}\\
    .\\
    .\\
    \color{magenta} u_1^{(n)} & \color{magenta} u_2^{(n)} ..... & \color{magenta} u_{99}^{(n)} & \color{magenta} u_{100}^{(n)}\\
    \color{magenta} .\\
    \color{magenta} u_1^{(n)} & \color{magenta} u_2^{(n)} ..... & \color{magenta} u_{99}^{(n)} & \color{magenta} u_{100}^{(n)}\\
    \color{magenta} u_1^{(n)} & \color{magenta} u_2^{(n)} ..... & \color{magenta} u_{99}^{(n)} & \color{magenta} u_{100}^{(n)}
    \end{bmatrix}_{nm\times100},\,\,\,\,\,\,\,\,\,\,
    \begin{bmatrix}
    \color{blue} y_1^{(1)}\\
    \color{blue} y_2^{(1)}\\
    .\\
    \color{blue} y_m^{(1)}\\
    \color{red} y_1^{(2)}\\
    \color{red} y_2^{(2)}\\
    \color{red} .\\
    \color{red} y_m^{(2)}\\
    .\\
    .\\
    \color{magenta} y_{(1)}^{(n)}\\
    \color{magenta} .\\
    \color{magenta} y_{(m-1)}^{(n)}\\
    \color{magenta} y_{(m)}^{(n)}
    \end{bmatrix}_{nm\times1},\,\,\,\,\,\,\,\,\,\,
    \begin{bmatrix}
    \color{blue} s_1^{(1)}\\
    \color{blue} s_2^{(1)}\\
    \color{blue} .\\
    \color{blue} s_m^{(1)}\\
    \color{red} s_1^{(2)}\\
    \color{red} s_2^{(2)}\\
    \color{red} .\\
    \color{red} s_m^{(2)}\\
    .\\
    .\\
    \color{magenta} s_{(1)}^{(n)}\\
    \color{magenta} .\\
    \color{magenta} s_{(m-1)}^{(n)}\\
    \color{magenta} s_{(m)}^{(n)}
    \end{bmatrix}_{nm\times1},
\end{equation}
where the subscript of $y$ is just referring to the index number of sample, $y_i$ however is selected randomly for each sample irrespective of input realization.
$s_i$ for a sample corresponds to $y_i$ and $\bm u$ of that particular sample.
The training data was normalized for the examples covering the ordinary differential equations.
The specific NN architecture selected for branch net and trunk net in the following examples is discussed in \ref{appendix: NN architectures}.
\subsection{Anti-derivative operator}
In this example, the nonlinear operator being learnt is an elementary differential operator defined as:
\begin{equation}
    \frac{ds(t)}{dt} = u(t),\,\,\,\,\,t\in[0,1]
\end{equation}
where $u(t)$ is the input function, $t$ is the independent variable and $s(t)$ is the target variable. 
Training data in this example comprises of 3000 unique input realizations and for each input realization, 20 time steps were selected at random for training.

Fig. \ref{figure: CI, individual realizations} shows the confidence intervals and the mean prediction corresponding to three different realizations of input force, obtained from the proposed VB-DeepONet framework.
It is observed that the proposed VB-DeepONet is not only able to capture the trend of the ground truth but also gives an uncertainty associated with its prediction.
It can also be seen that the VB-DeepONet is not just learning a constant uncertainty bound but rather has the capacity to judge the uncertainty associated with predictions corresponding to different inputs.
\begin{figure}[ht!]
\centering
\begin{subfigure}{0.45\textwidth}
\includegraphics[width = \textwidth]{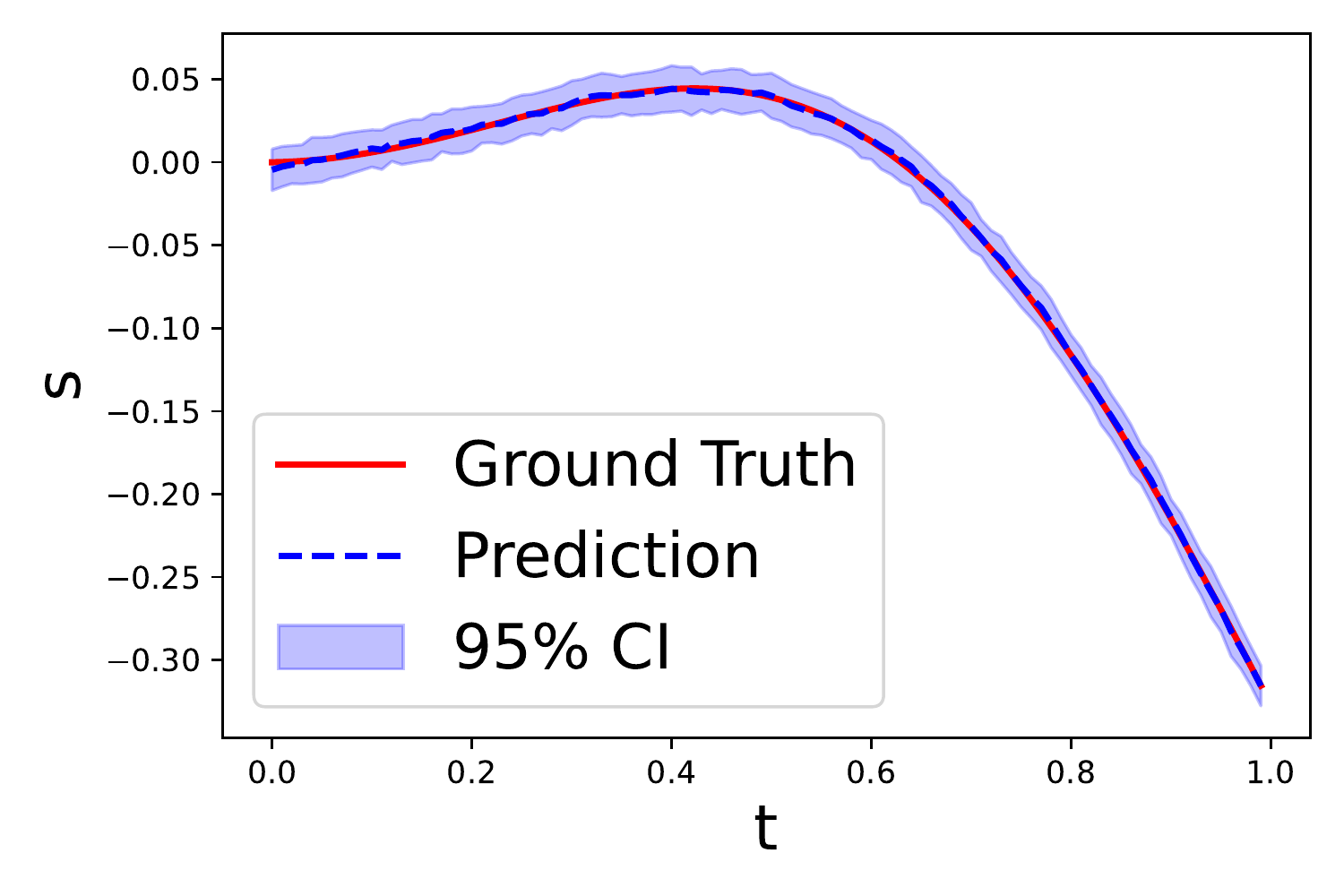}
\caption{Sample \#1}
\end{subfigure}
\begin{subfigure}{0.45\textwidth}
\includegraphics[width = \textwidth]{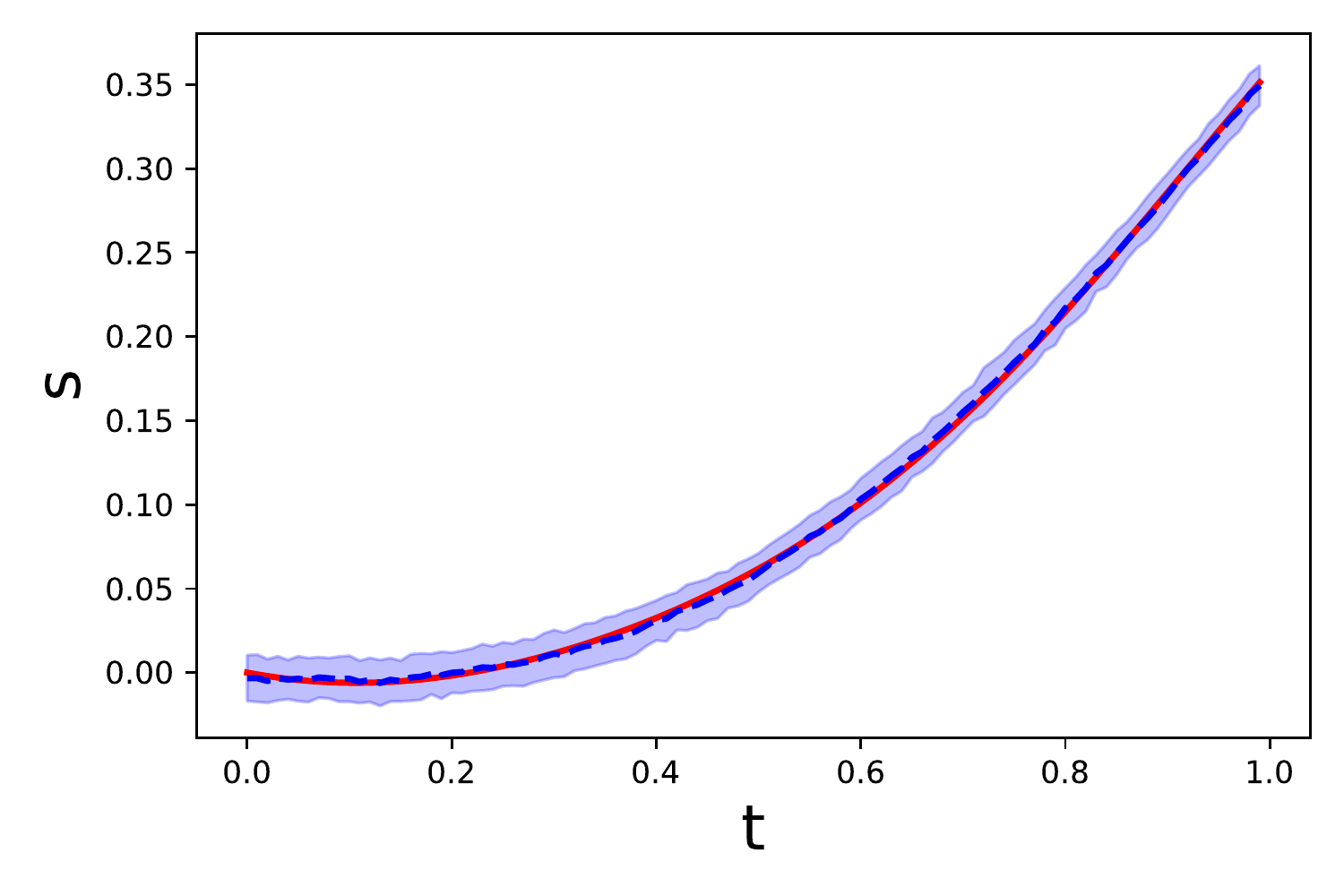}
\caption{Sample \#2}
\end{subfigure}
\begin{subfigure}{0.45\textwidth}
\includegraphics[width = \textwidth]{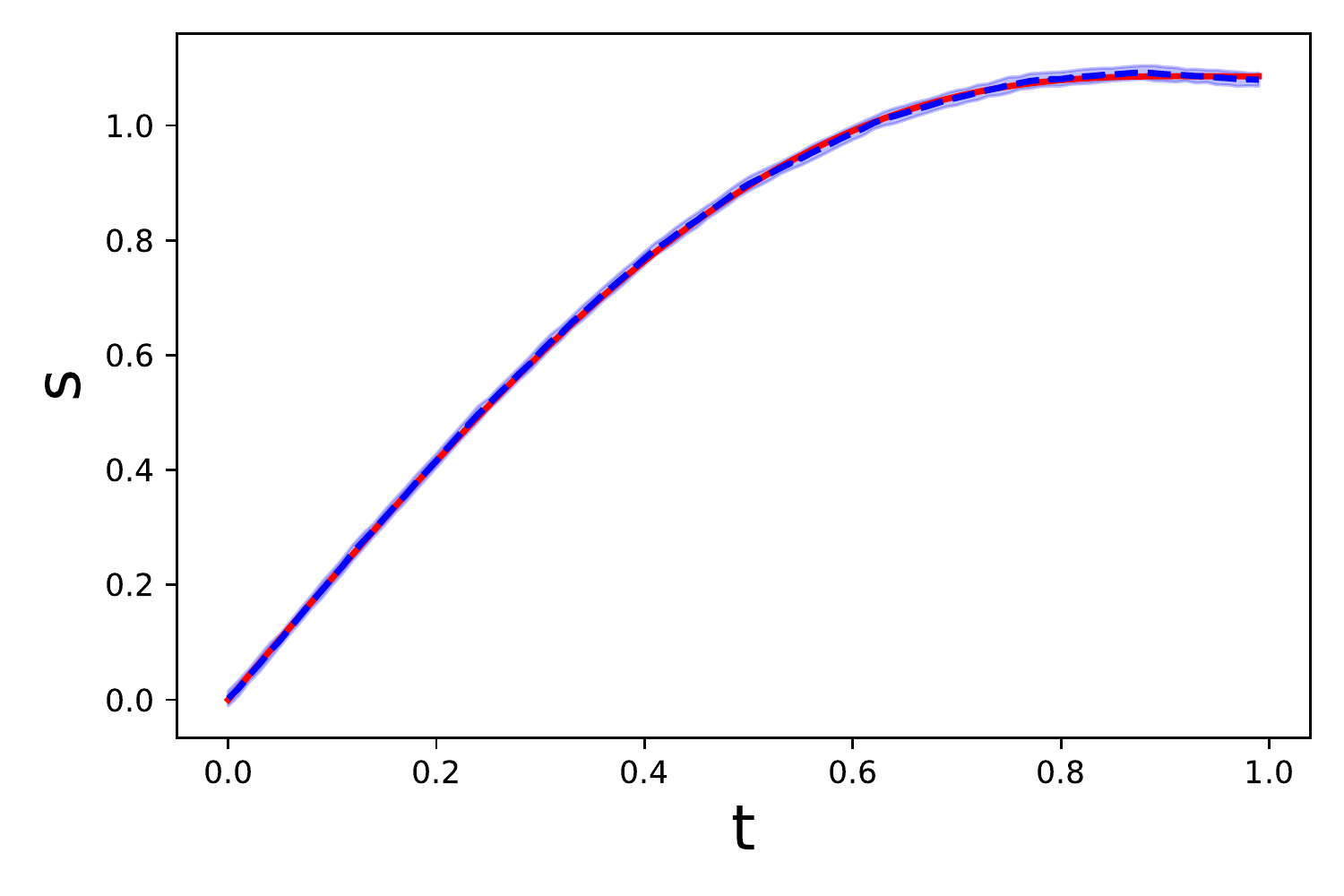}
\caption{Sample \#3}
\end{subfigure}
\caption{VB-DeepONet predictions corresponding to three different testing input realizations compared against the ground truth for AD example.}
\label{figure: CI, individual realizations}
\end{figure}

The trained model was used for quantifying output uncertainty because of the aleatoric uncertainty associated with the system.
To that end, 10000 input realizations are generated and the complete time series corresponding to each input realization, discretized at 100 uniformly spaced time steps $t_{1:100}$ was used to generate testing samples.
Probability density functions (PDF), for predicted target solution $s$ at $t_7$ and $t_{10}$ are plotted in Fig. \ref{figure: CI, pdf}.
As can be seen that the mean PDF, follows the ground truth closely and even at the places where the mean PDF slightly deviates from the ground truth (GT), the GT is completely engulfed by the confidence intervals for the PDF.
Comparison has also been drawn against D-DeepONet results and while it closely follows the ground truth, it is comparatively less accurate as compared to the proposed VB-DeepONet. Also, there are no confidence intervals for the D-DeepONet results; this may create inconvenience in taking potentially risky decisions.
\begin{figure}[ht!]
\centering
\begin{subfigure}{0.75\textwidth}
\includegraphics[width = \textwidth]{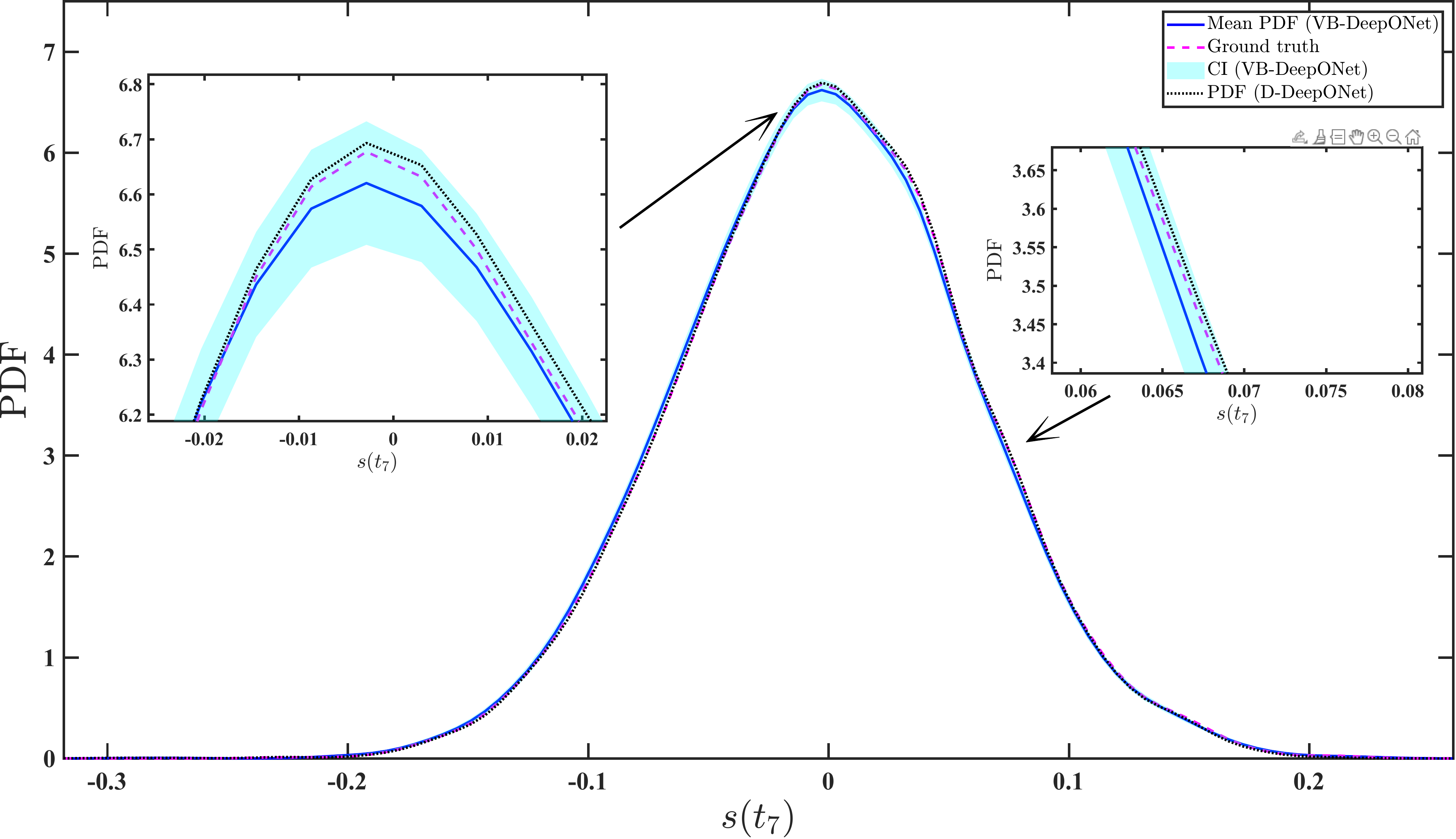}
\caption{PDF, s($t_7$)}
\end{subfigure}
\begin{subfigure}{0.75\textwidth}
\includegraphics[width = \textwidth]{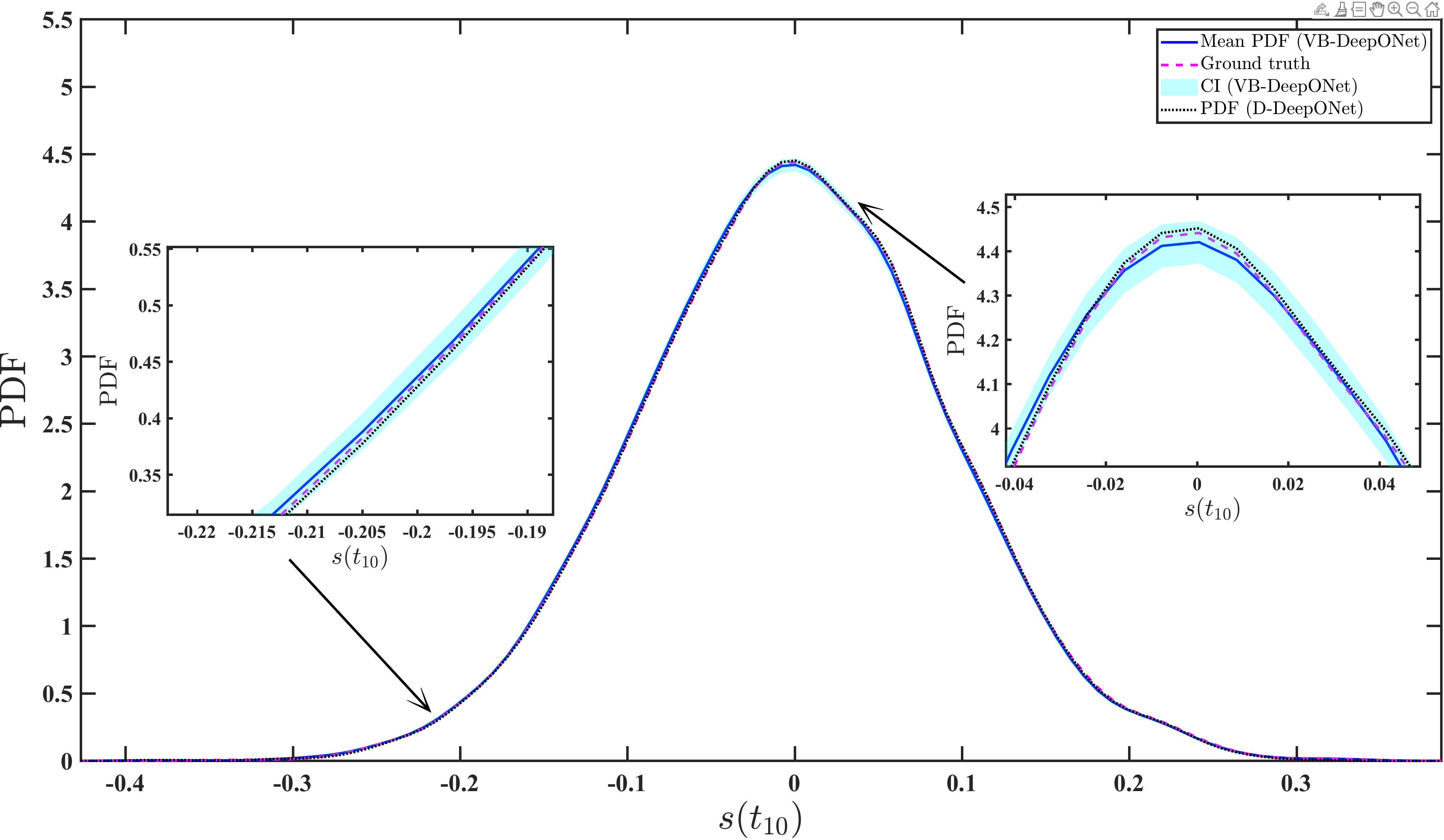}
\caption{PDF, s($t_{10}$)}
\end{subfigure}
\caption{PDFs plotted at time step $t_7$ and $t_{10}$ for the AD example.}
\label{figure: CI, pdf}
\end{figure}

To further illustrate the robustness of the VB-DeepONet model, it was trained for different number of unique initial realizations and the NMSE obtained during testing stage are plotted in Fig. \ref{figure: CI, comp training samples}.
The evolution of NMSE follows the expected trend i.e., as the number of training samples increase, the testing error decreases.
\begin{figure}[ht!]
\centering
\includegraphics[width = 0.85\textwidth]{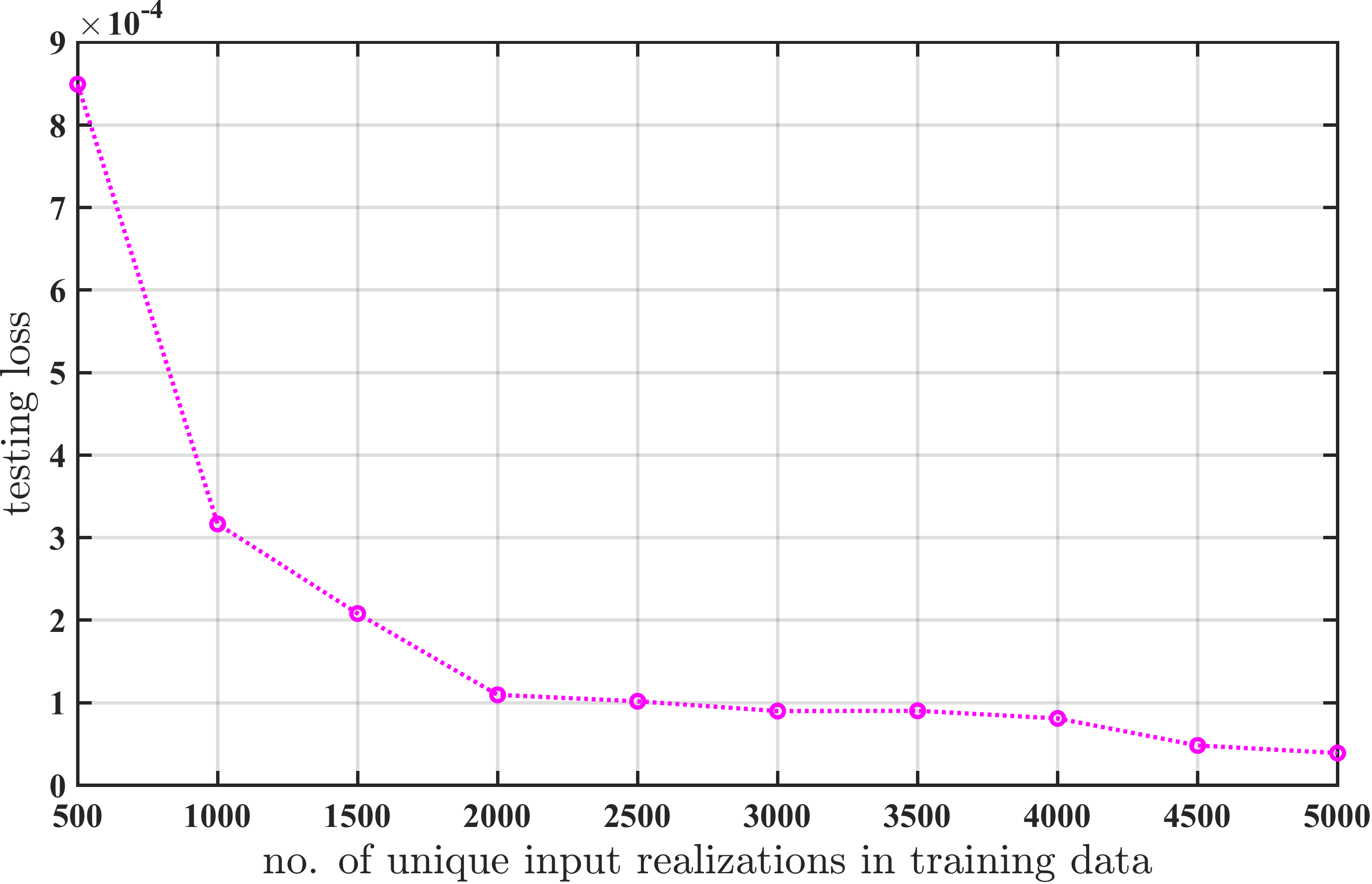}
\caption{Evolution of testing loss with increase in training data for the AD example.}
\label{figure: CI, comp training samples}
\end{figure}
\subsection{Gravity pendulum subjected to external force}
A gravity pendulum is explored in this example, governing equation for which is defined as follows:
\begin{equation}
    \begin{gathered}
    \dfrac{ds_1}{dt} = s_2\\
    \dfrac{ds_2}{dt} = -sin(s_1)+u(t),
    \end{gathered}
\end{equation}
where $t$ is the independent variable and $u(t
)$ is the input function with $s_1(t)$  as the target output.
The training data consists of 3500 unique input forces with 20 random time steps selected for each input.

\begin{figure}[ht!]
\centering
\begin{subfigure}{0.45\textwidth}
\includegraphics[width = \textwidth]{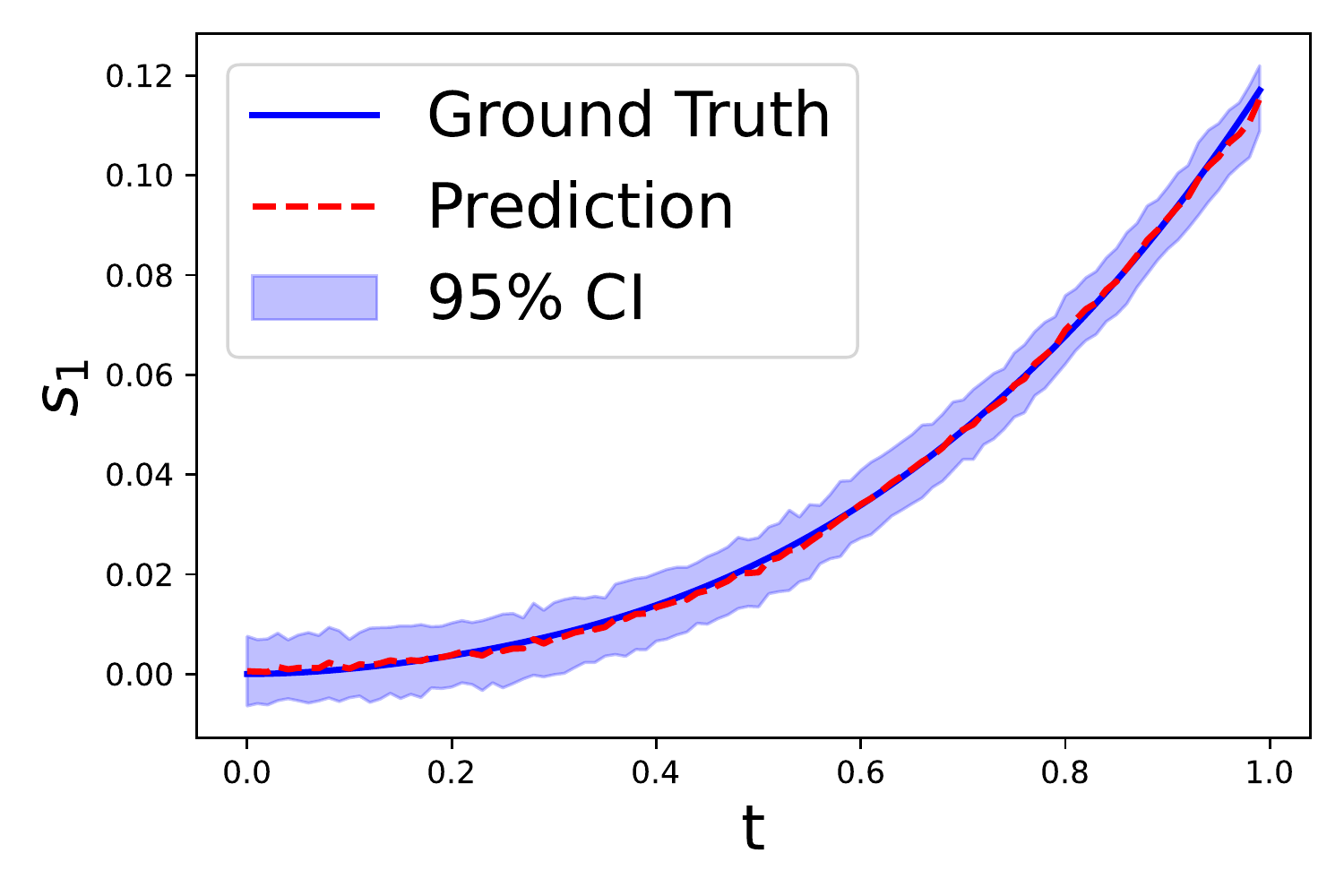}
\caption{Sample \#1}
\end{subfigure}
\begin{subfigure}{0.45\textwidth}
\includegraphics[width = \textwidth]{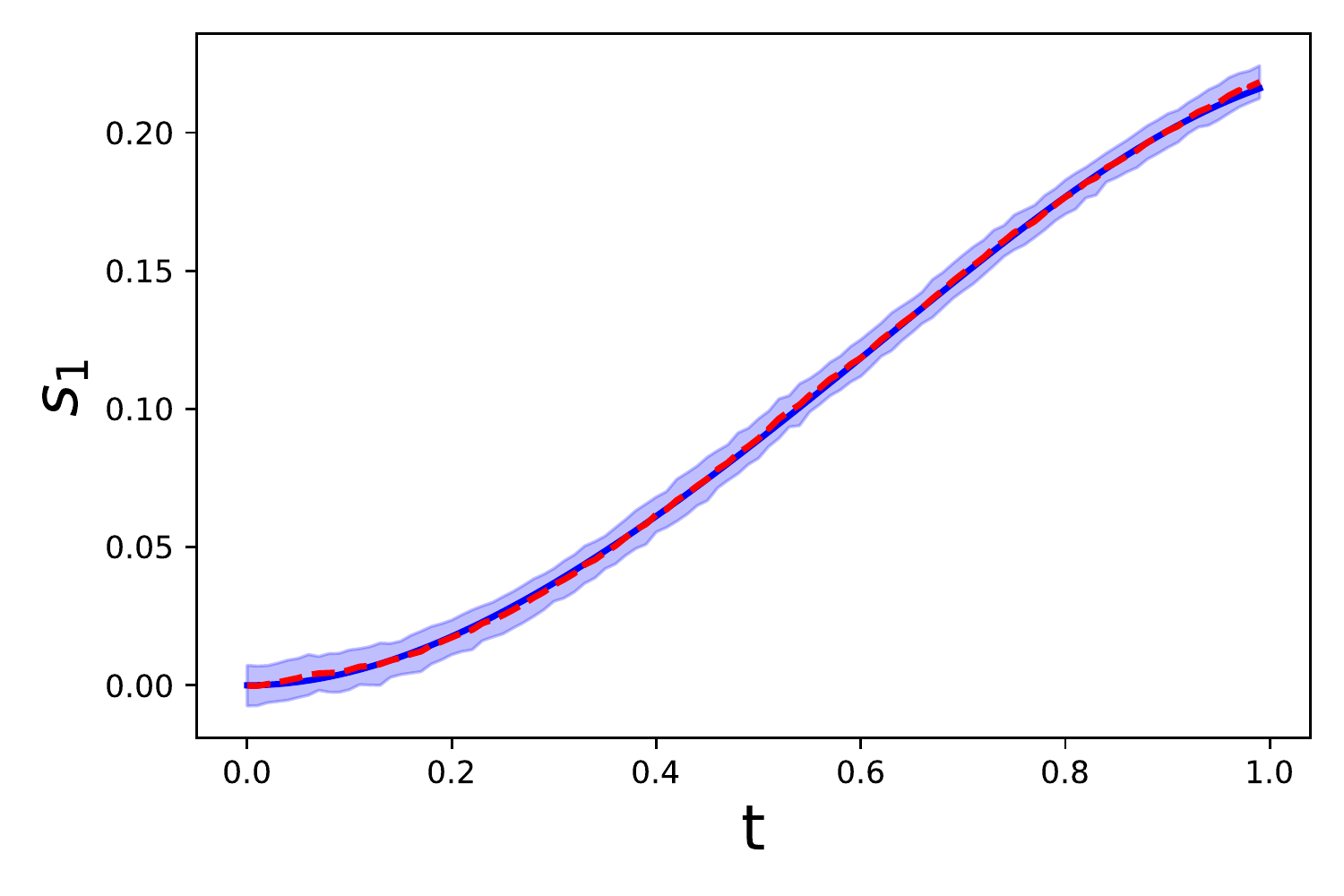}
\caption{Sample \#2}
\end{subfigure}
\begin{subfigure}{0.45\textwidth}
\includegraphics[width = \textwidth]{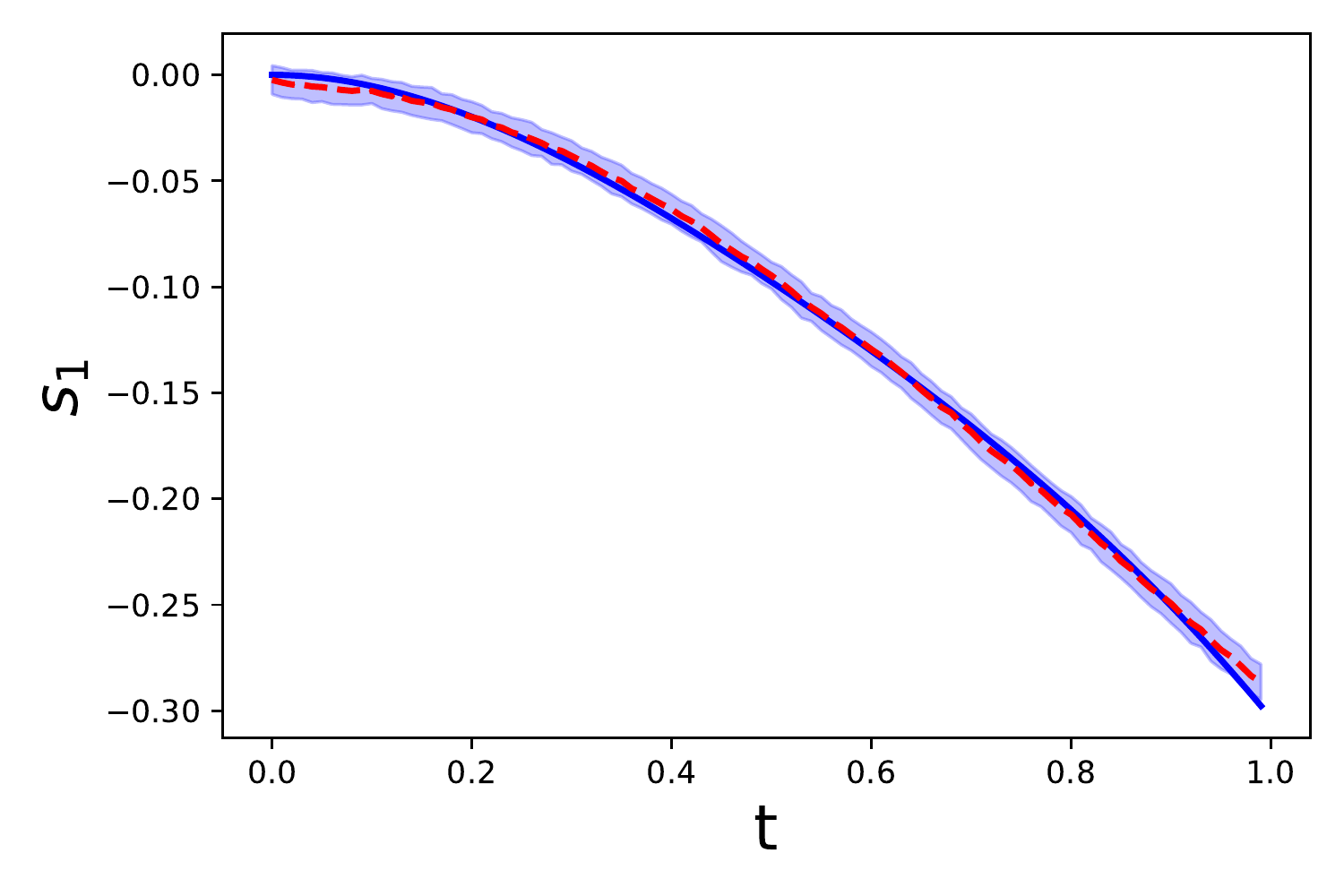}
\caption{Sample \#3}
\end{subfigure}
\caption{VB-DeepONet predictions compared against the ground truth for the gravity pendulum example, corresponding to three different testing input realizations.}
\label{figure: CII without noise}
\end{figure}

Fig. \ref{figure: CII without noise} shows VB-DeepONet results compared against the ground truth corresponding to three different input realizations.
As observed in the previous example, the VB-DeepONet is not just learning a constant standard deviation for all its predictions but is rather capable of estimating the uncertainty associated with different predictions.
As for the mean prediction, they follow the ground truth closely and give a good approximation of the same.
Similar to previous example, to estimate the output aleatoric uncertainty, 10000 unique input realizations were generated and the complete time series were used to generate testing samples corresponding to each input realization.
PDFs for target solution $s$ at $t_{30}$ and $t_{50}$ are plotted in Fig. \ref{figure: CII, pdf}.
The mean PDF from VB-DeepONet results again closely follows the ground truth while the confidence interval manages to account for small ambiguities between the mean PDF and the GT.
Similar to previous example, while D-DeepONet closely follows the ground truth, lack of confidence intervals might hinder the users ability to take well informed decisions. Performance wise, the VB-DeepONet predicted PDF plot have a slightly better match as compared to the D-DeepONet predicted PDF.
\begin{figure}[ht!]
\centering
\begin{subfigure}{0.75\textwidth}
\includegraphics[width = \textwidth]{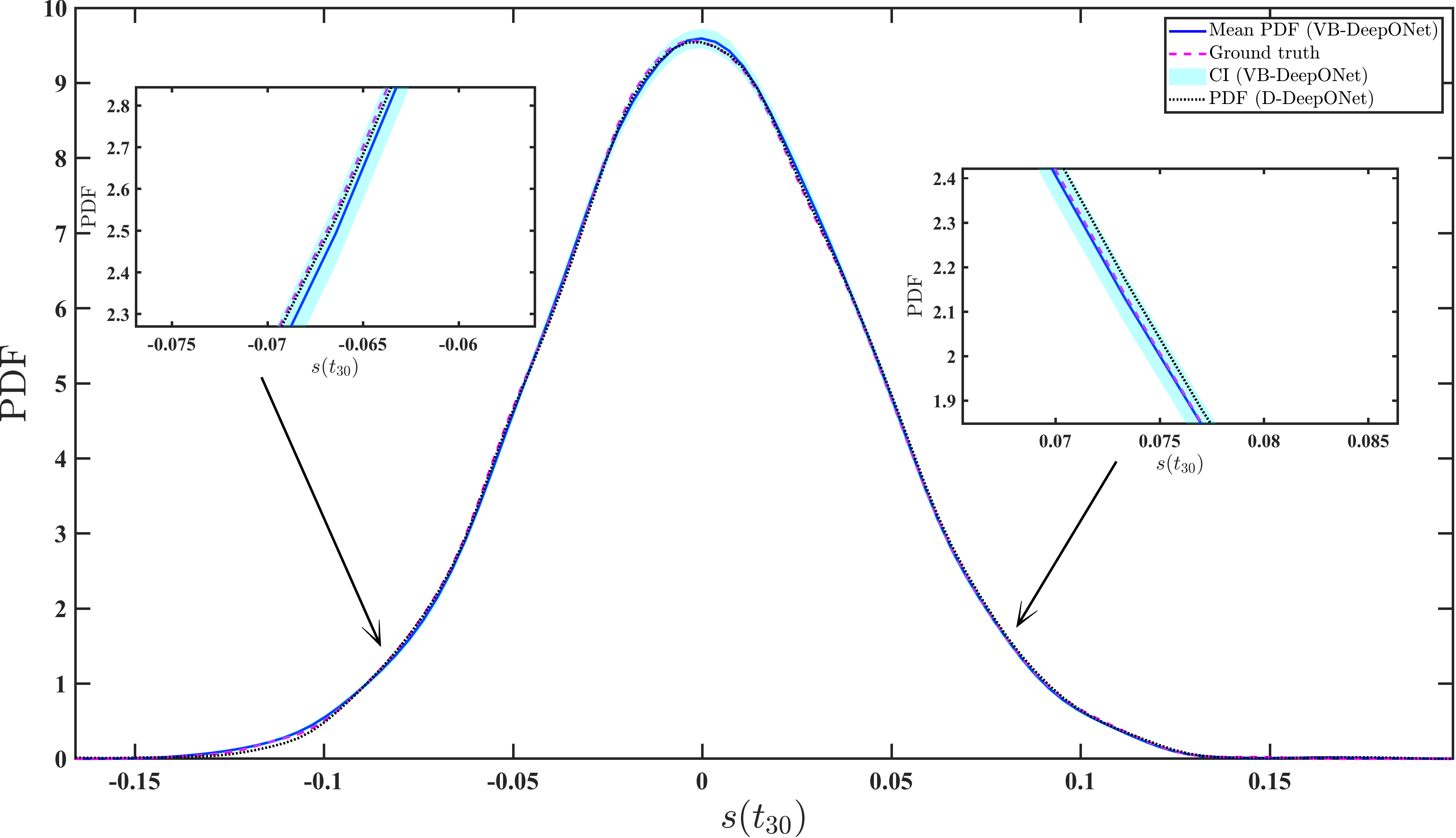}
\caption{PDF, s($t_{30}$)}
\end{subfigure}
\begin{subfigure}{0.75\textwidth}
\includegraphics[width = \textwidth]{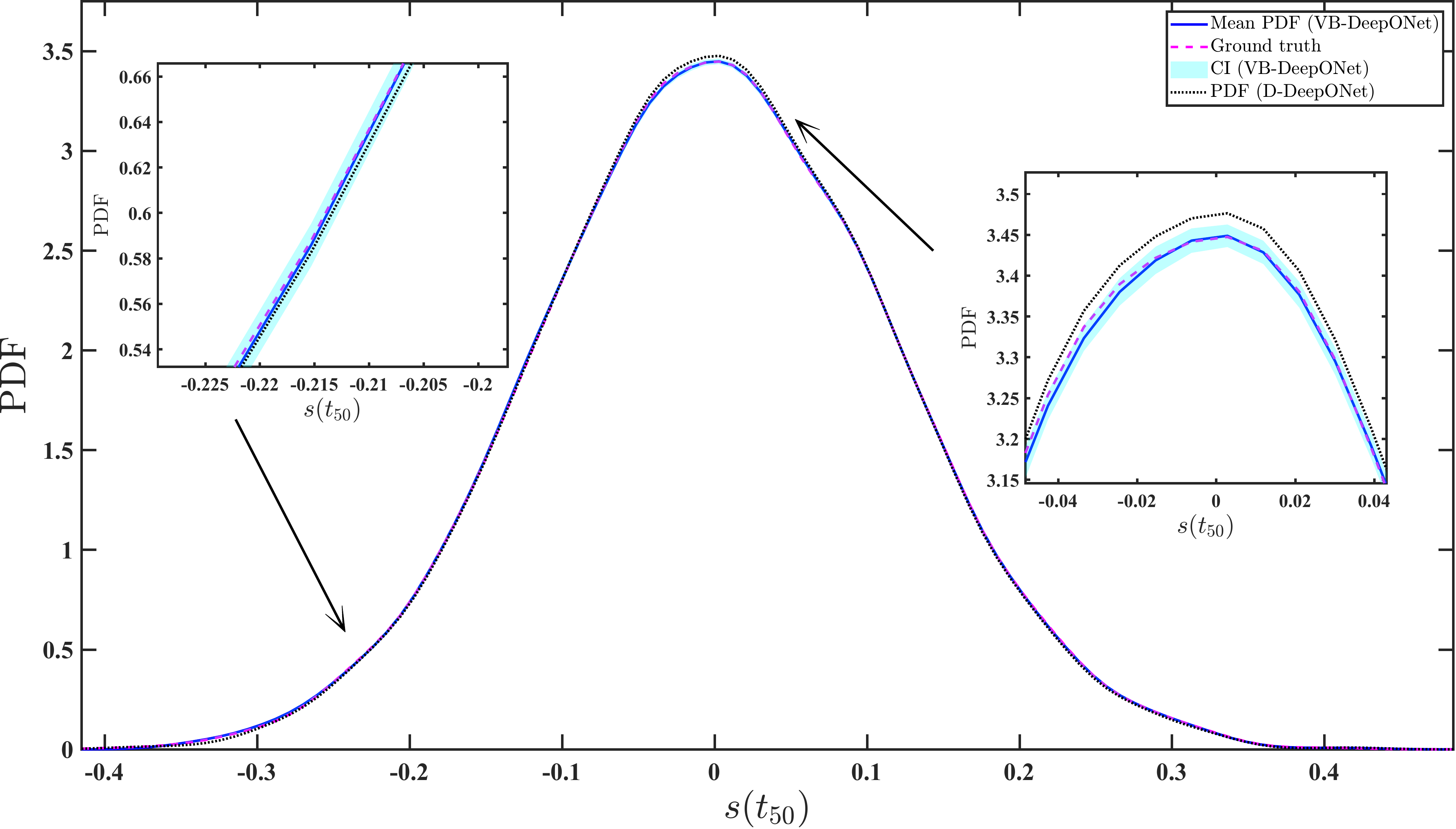}
\caption{PDF, s($t_{50}$)}
\end{subfigure}
\caption{PDFs plotted at time step $t_{30}$ and $t_{50}$ for the gravity pendulum example.}
\label{figure: CII, pdf}
\end{figure}
\subsection{Diffusion reaction system}
This example deals with a partial DE having fractional derivatives in both time and space.
It represents a Diffusion Reaction (DR) system with zero initial and boundary conditions and is defined as,
\begin{equation}
    \frac{\partial s}{\partial t} = D\frac{\partial^2s}{\partial x^2}+ks^2+u(x),\,\,\,\,\,x\in[0,1],\,\,t\in[0,1],
\end{equation}
where $x$ and $t$ are the independent variables, $u(x)$ is the input function and $s(x,t)$  is the target output.
The training and testing data for this example is generated using the GitHub codes provided for the paper \cite{lu2021learning}.
The training data consists of 500 unique input forces with 100 random sets of $x$ and $t$ selected for each unique input realization.
Table \ref{Table: CIII} shows solution corresponding to one realization of the testing input.
The mean predictions for this example closely matches with the ground truth and the same is confirmed from the absolute error computed with respect to ground truth.
Standard deviation observed using VB-DeepONet is also plotted, which details the uncertainty associated with predictions at different grid points (locations of $x$ and $t$).
\begin{table}[ht!]
\begin{tabular}{cccc}
\hline Algorithm & Solution & Absolute Error & Std. dev. \\\hline\\

Ground Truth &
\begin{subfigure}{0.245\textwidth}
\centering
\includegraphics[width = \textwidth]{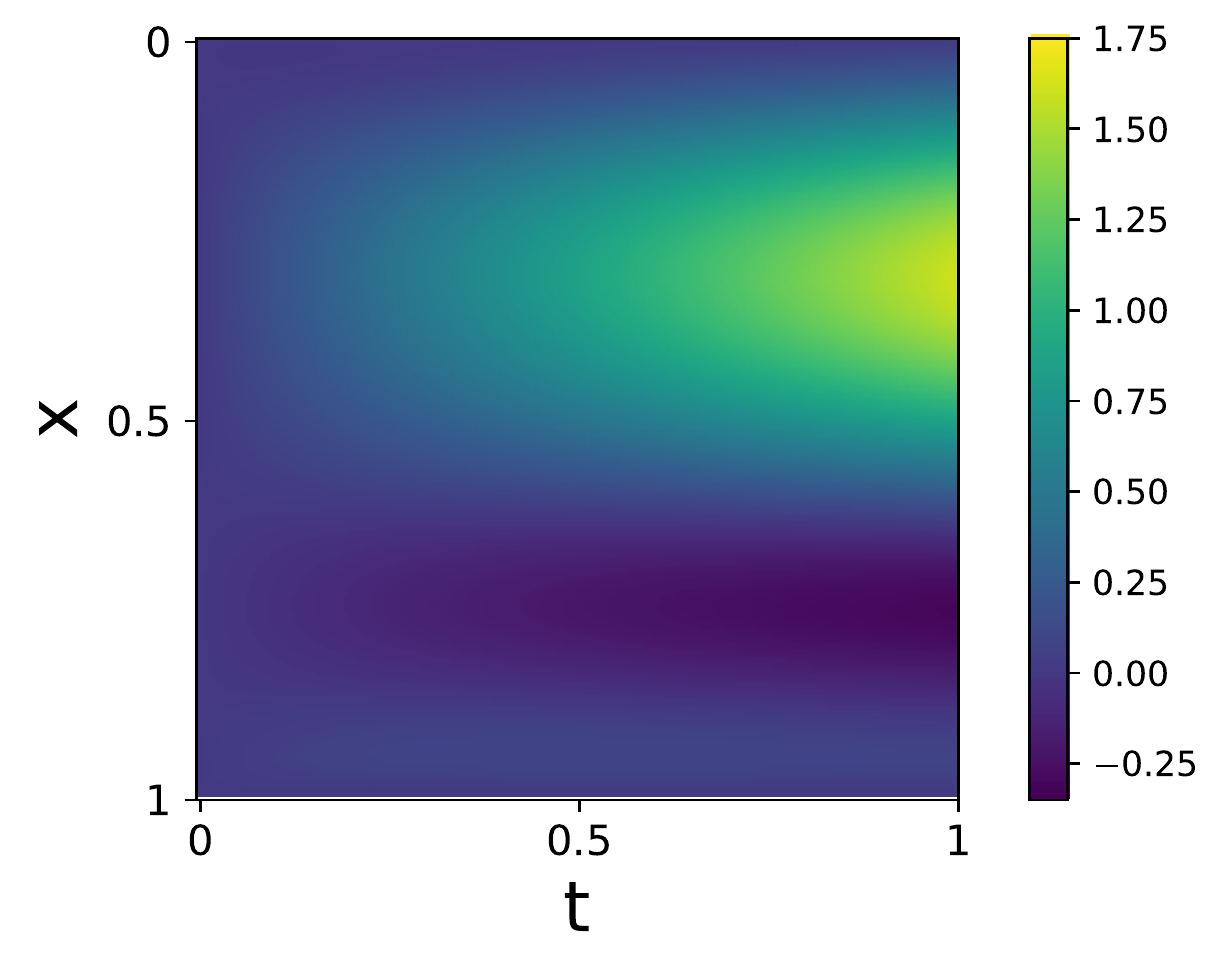}
\end{subfigure} &  
 - & 
 - \\[12pt]

D-DeepONet &
\begin{subfigure}{0.245\textwidth}
\includegraphics[width = \textwidth]{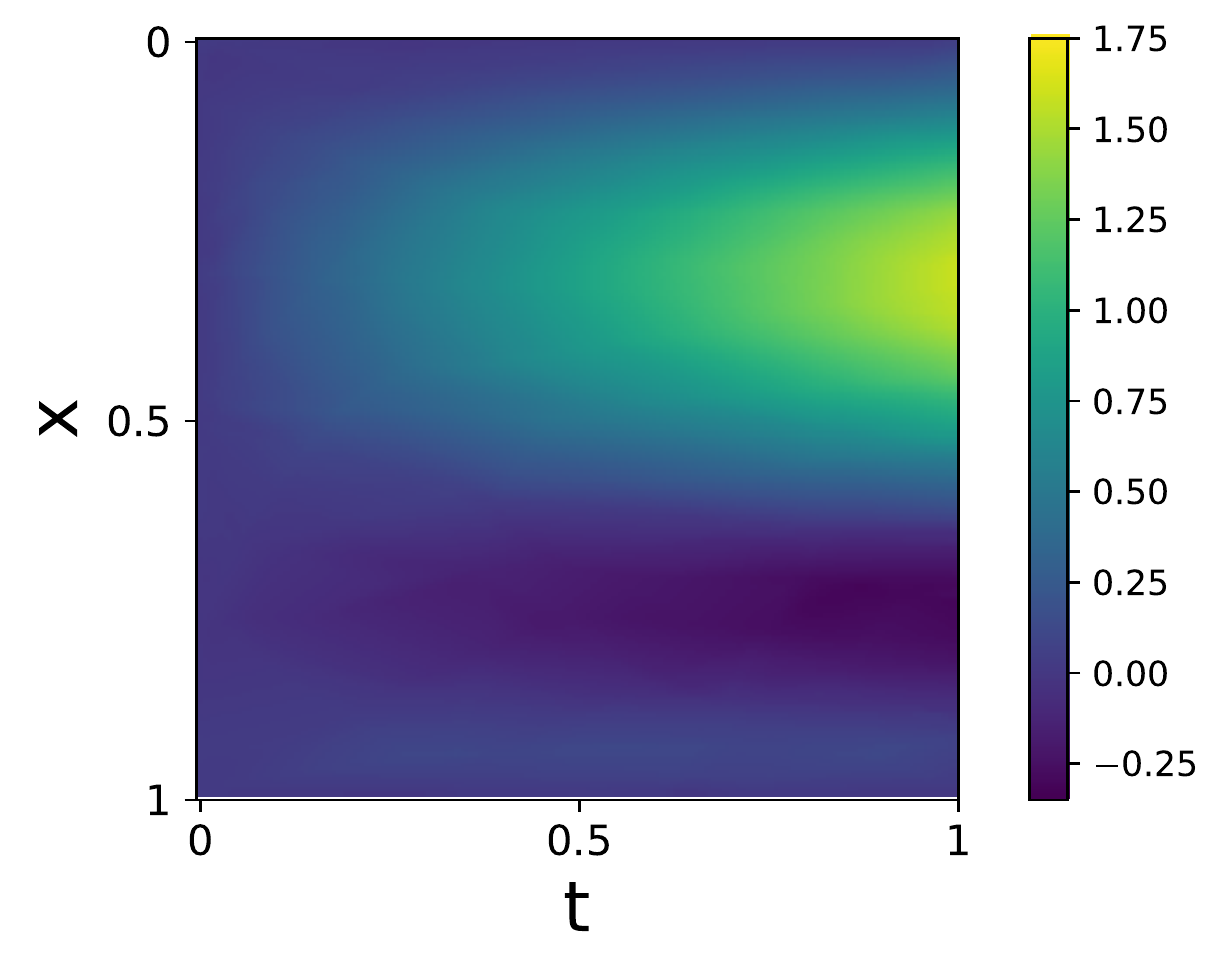}
\end{subfigure} &
\begin{subfigure}{0.245\textwidth}
\includegraphics[width = \textwidth]{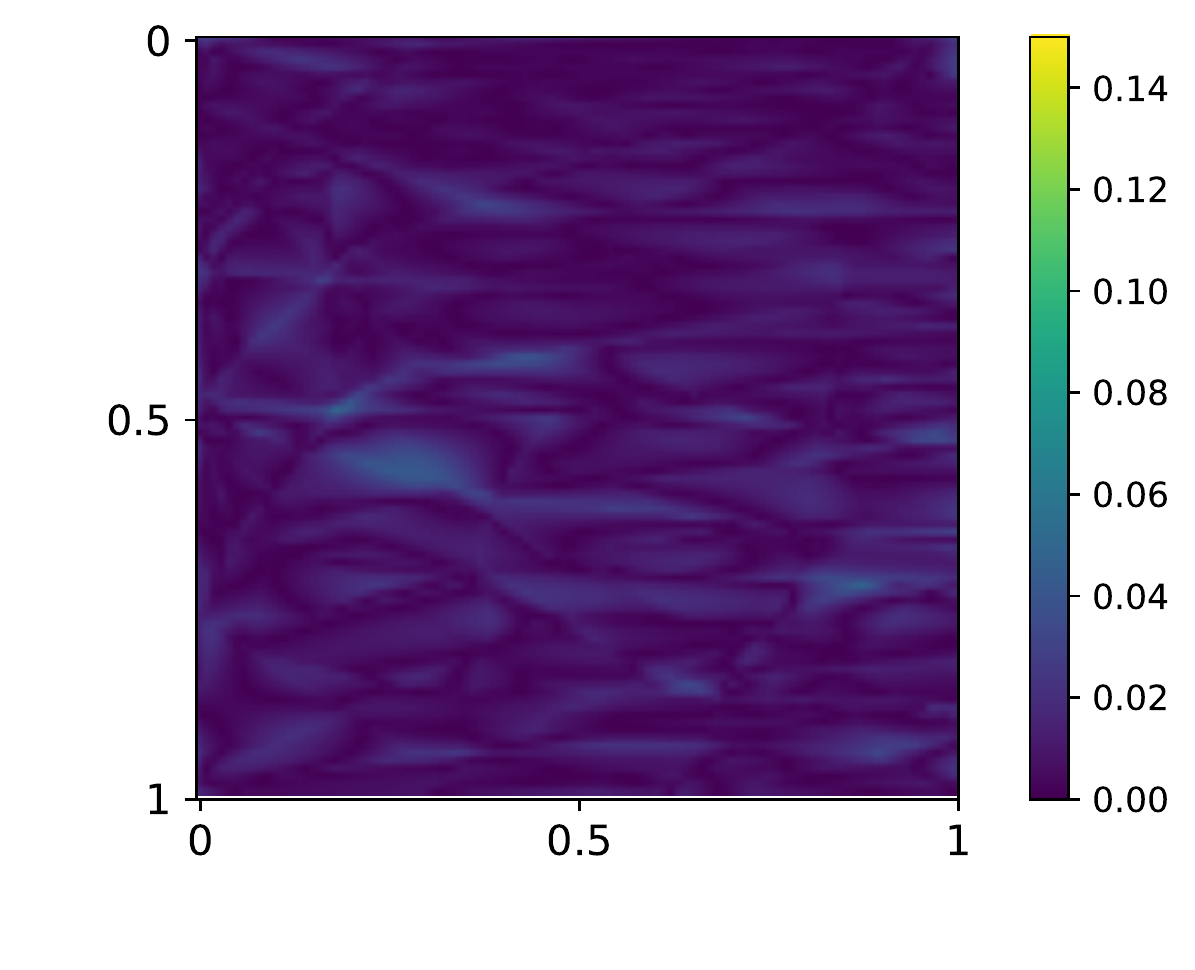}
\end{subfigure} &
\textbf{ - }\\[12pt]

VB-DeepONet &
\begin{subfigure}{0.245\textwidth}
\includegraphics[width = \textwidth]{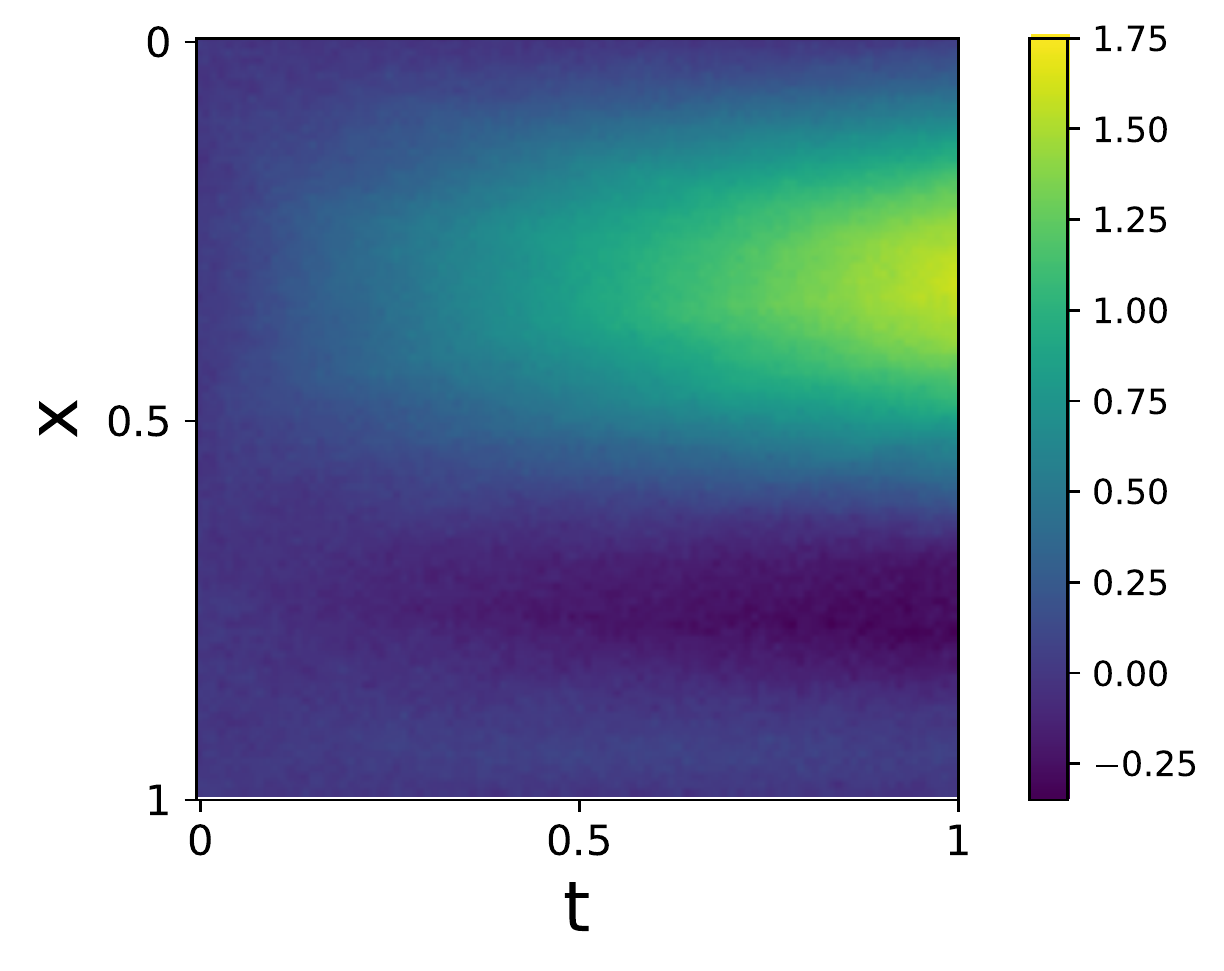}
\end{subfigure} &
\begin{subfigure}{0.245\textwidth}
\includegraphics[width = \textwidth]{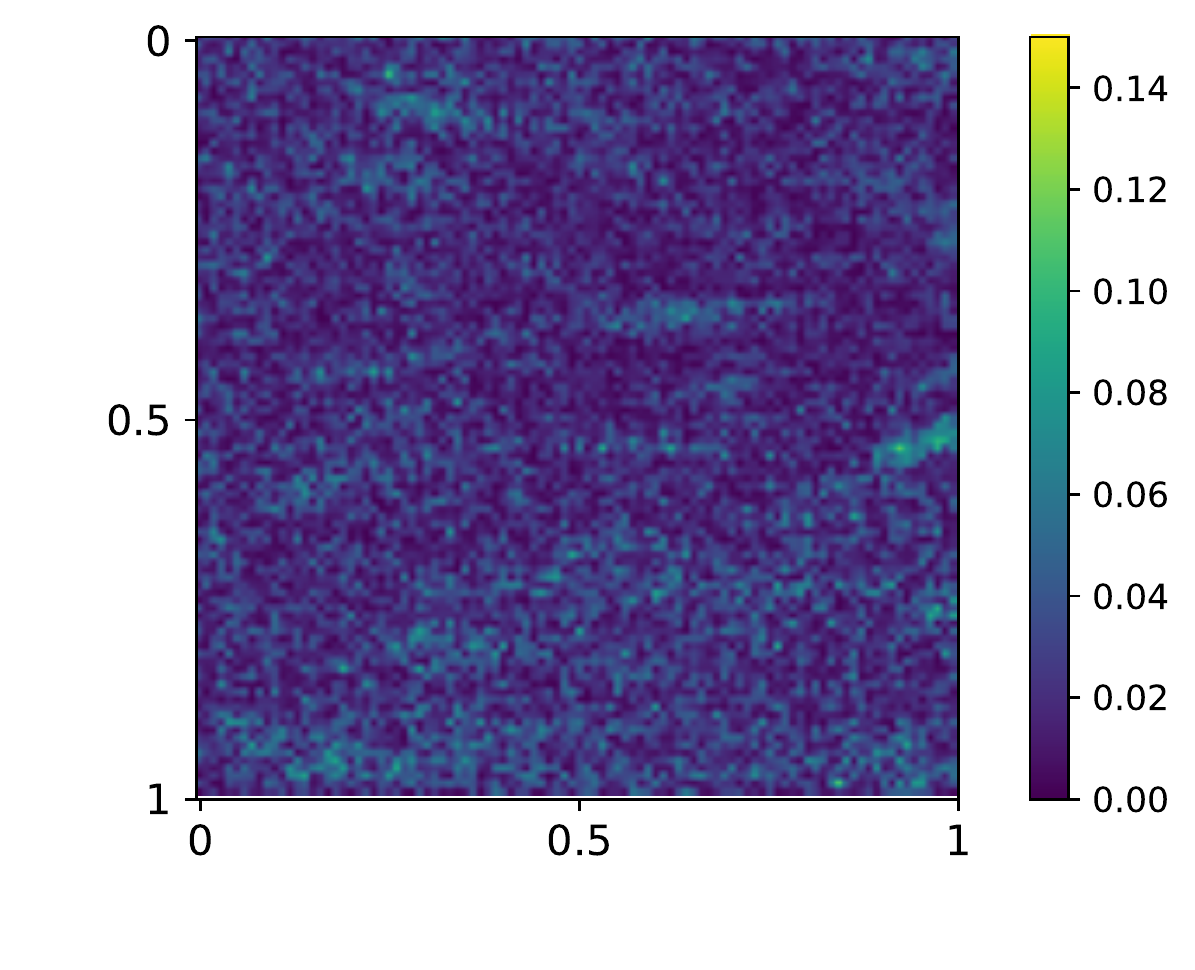}
\end{subfigure} &
\begin{subfigure}{0.245\textwidth}
\includegraphics[width = \textwidth]{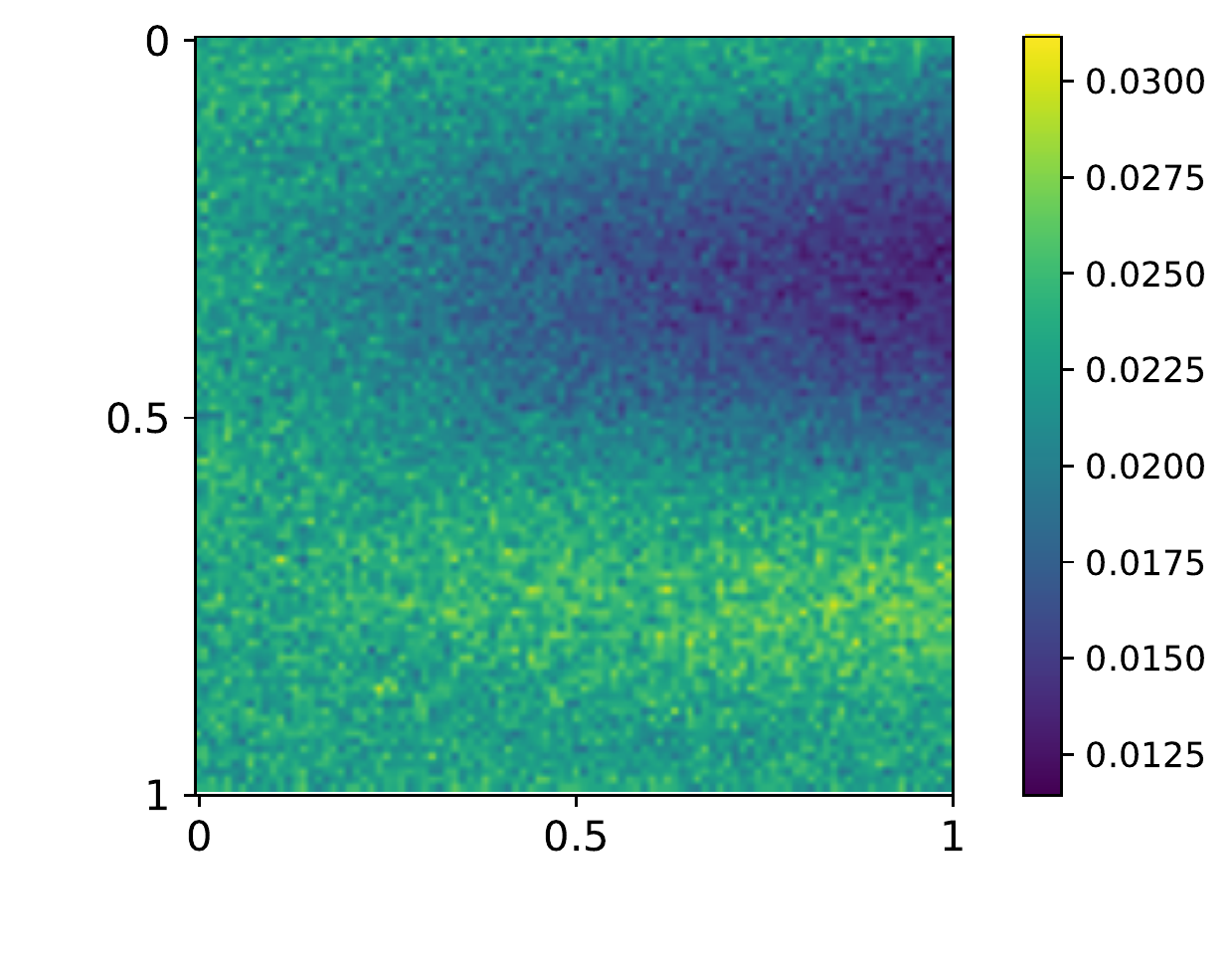}
\end{subfigure}
\\\hline
\end{tabular}
\vspace{0.25cm}
\caption{VB-DeepONet and D-DeepONet predictions compared against the ground truth for a single testing input realization, for DR example.}
\label{Table: CIII}
\end{table}

In Fig. \ref{figure: CIII, CIs}, confidence intervals along with the mean prediction for the VB-DeepONet are plotted and it can be observed that the limits of the confidence interval are not arbitrarily large but rather give a good representation of the ground truth themselves.
\begin{figure}[ht!]
\centering
\begin{subfigure}{0.32\textwidth}
\includegraphics[width = \textwidth]{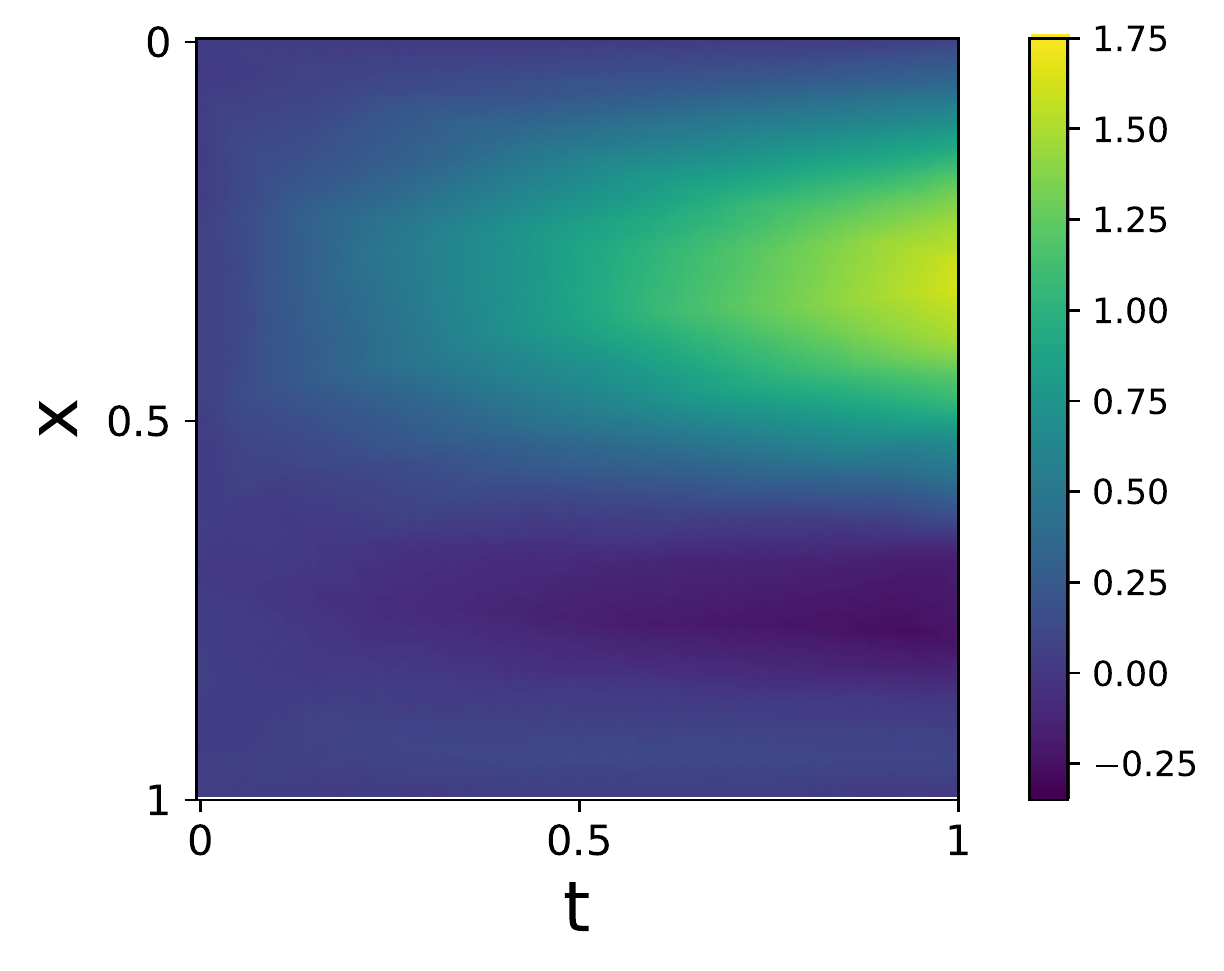}
\caption{\centering VB-DeepONet, 95\% CI, Upper Limit}
\end{subfigure}
\begin{subfigure}{0.32\textwidth}
\includegraphics[width = \textwidth]{DR_BDN_PRED.pdf}
\caption{VB-DeepONet, Mean Prediction}
\end{subfigure}
\begin{subfigure}{0.32\textwidth}
\includegraphics[width = \textwidth]{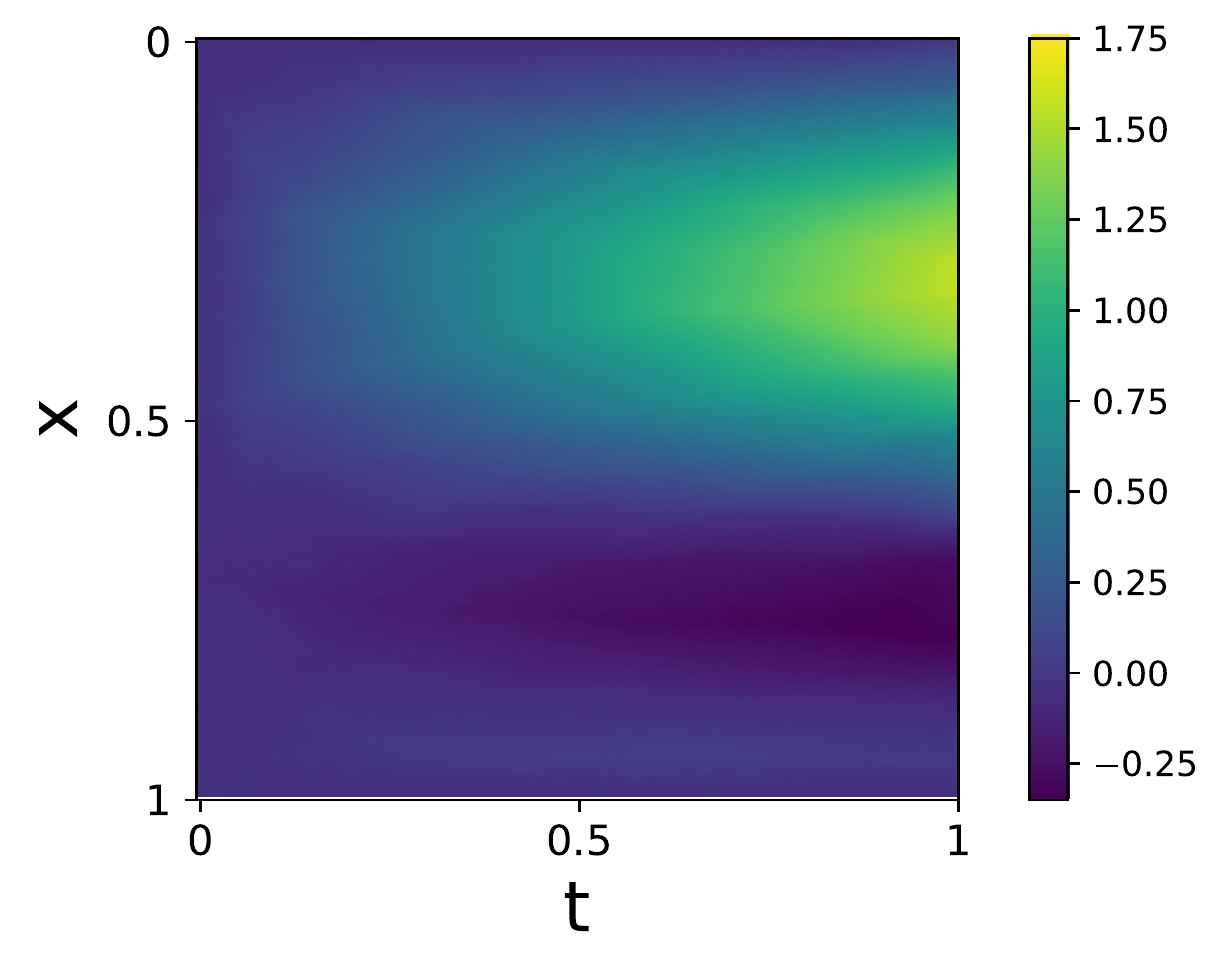}
\caption{\centering VB-DeepONet, 95\% CI, Lower Limit}
\end{subfigure}
\caption{VB-DeepONet results corresponding to single input realization for DR example.}
\label{figure: CIII, CIs}
\end{figure}
The PDFs plotted in Fig. \ref{figure: CIII, PDFs} corresponds to input data having 10000 unique input realizations and for each input realization, solutions corresponding to $100\times100$ space-time grid were used to generate the samples.
As can be seen that in this example, D-DeepONet results have slight discrepancy as compared to the ground truth and the lack of confidence intervals further aggravate the discrepancy observed.
The VB-DeepONet however closely follows the ground truth with the associated uncertainty shown via the confidence intervals.
\begin{figure}[ht!]
\centering
\begin{subfigure}{0.75\textwidth}
\includegraphics[width = \textwidth]{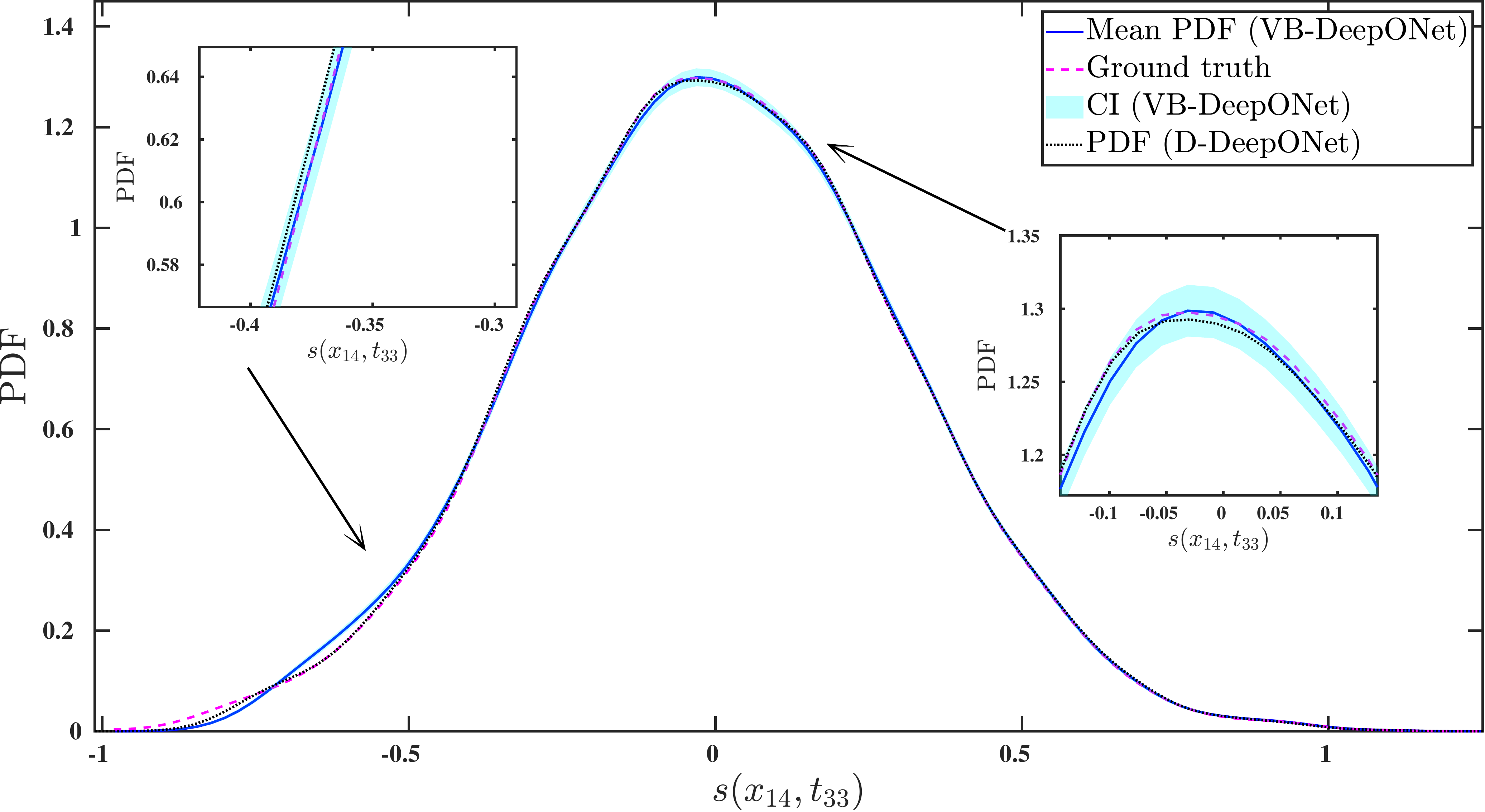}
\caption{PDF, $s(x_8,t_{45})$}
\end{subfigure}
\begin{subfigure}{0.75\textwidth}
\includegraphics[width = \textwidth]{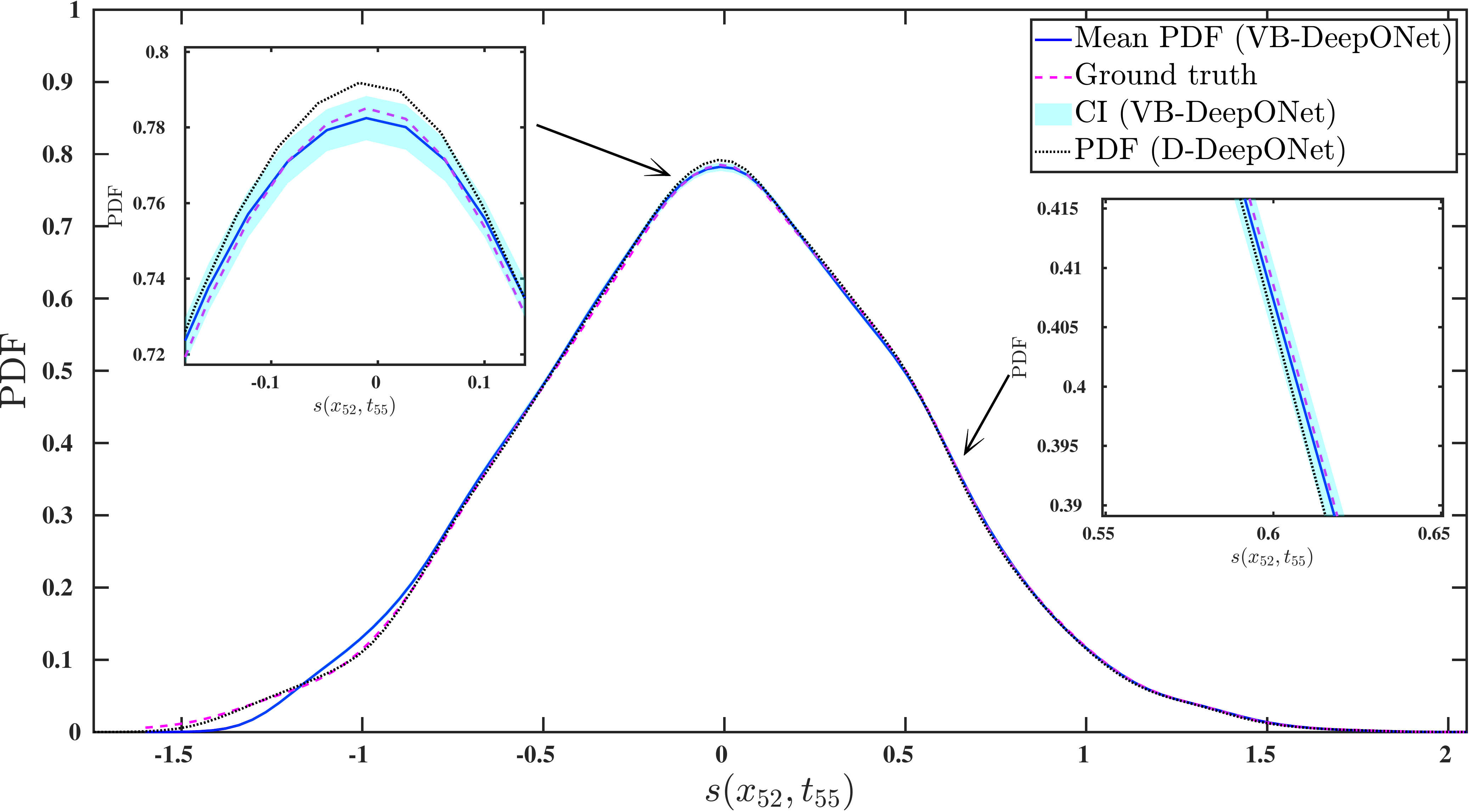}
\caption{PDF, $s(x_{47},t_{100})$}
\end{subfigure}
\caption{PDFs at different pairs of $x$ and $t$ for DR system.}
\label{figure: CIII, PDFs}
\end{figure}
\subsection{Advection-Diffusion equation}
In this last example, an Advection Diffusion (ADVD) equation with periodic boundary conditions is learnt using the proposed VB-DeepONet.
Similar to previous example, this PDE also has fractional derivatives in both time and space, and is defined as: 
\begin{equation}
    \frac{\partial s}{\partial t}+\frac{\partial s}{\partial x}-0.1\frac{\partial^2s}{\partial x^2}=0,\,\,\,\,\,s(x,0) = sin^2(2
    \pi u(x)),\,\,x\in[0,1],\,\,t\in[0,1],
\end{equation}
The challenge in this example however is that the partial DE under consideration is parametric by virtue of its changing initial condition and hence the mapping is done between the initial condition $s(x,0)$ and the final solution $s(x,t)$.
Data for this example is generated using the GitHub codes for the paper \cite{lu2021learning}.
Data corresponding to 1000 unique input forces was selected for training, with 100 random sets of $x$ and $t$ selected for each unique input realization.

Table \ref{table: CIV} shows the result corresponding to one realization of testing input along with the absolute error in VB-DeepONet and D-DeepONet predictions when compared against the ground truth.
The results produced shows excellent generalization capabilities of the VB-DeepONet, along with the capacity to quantify uncertainty associated with its predictions.
\begin{table}[ht!]
\begin{tabular}{cccc}
\hline Algorithm & Solution & Absolute Error & Std. dev. \\\hline\\
 
Ground Truth &
\begin{subfigure}{0.245\textwidth}
\centering
\includegraphics[width = \textwidth]{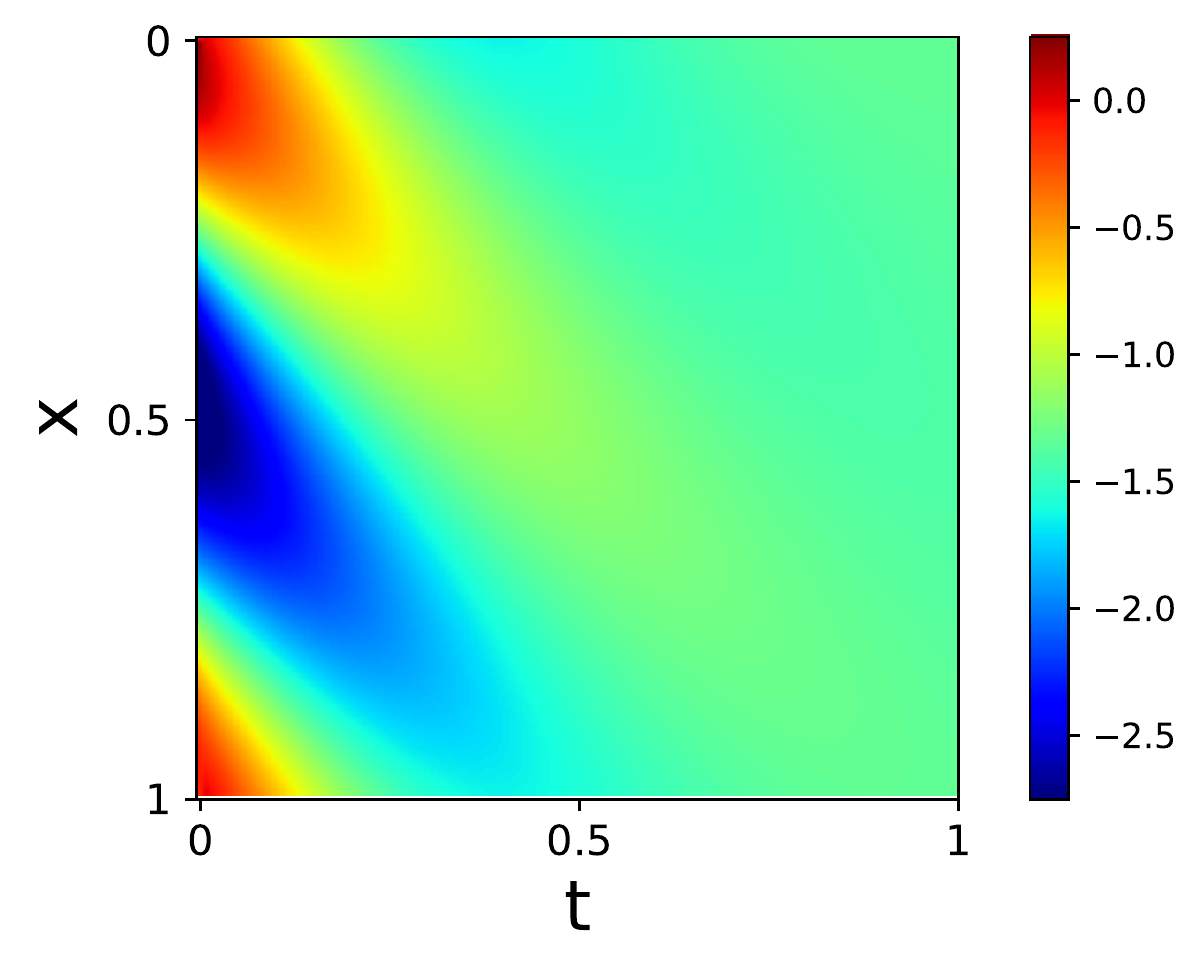}
\end{subfigure} &  
\textbf{ - }& 
\textbf{ - }\\[12pt]

D-DeepONet &
\begin{subfigure}{0.245\textwidth}
\includegraphics[width = \textwidth]{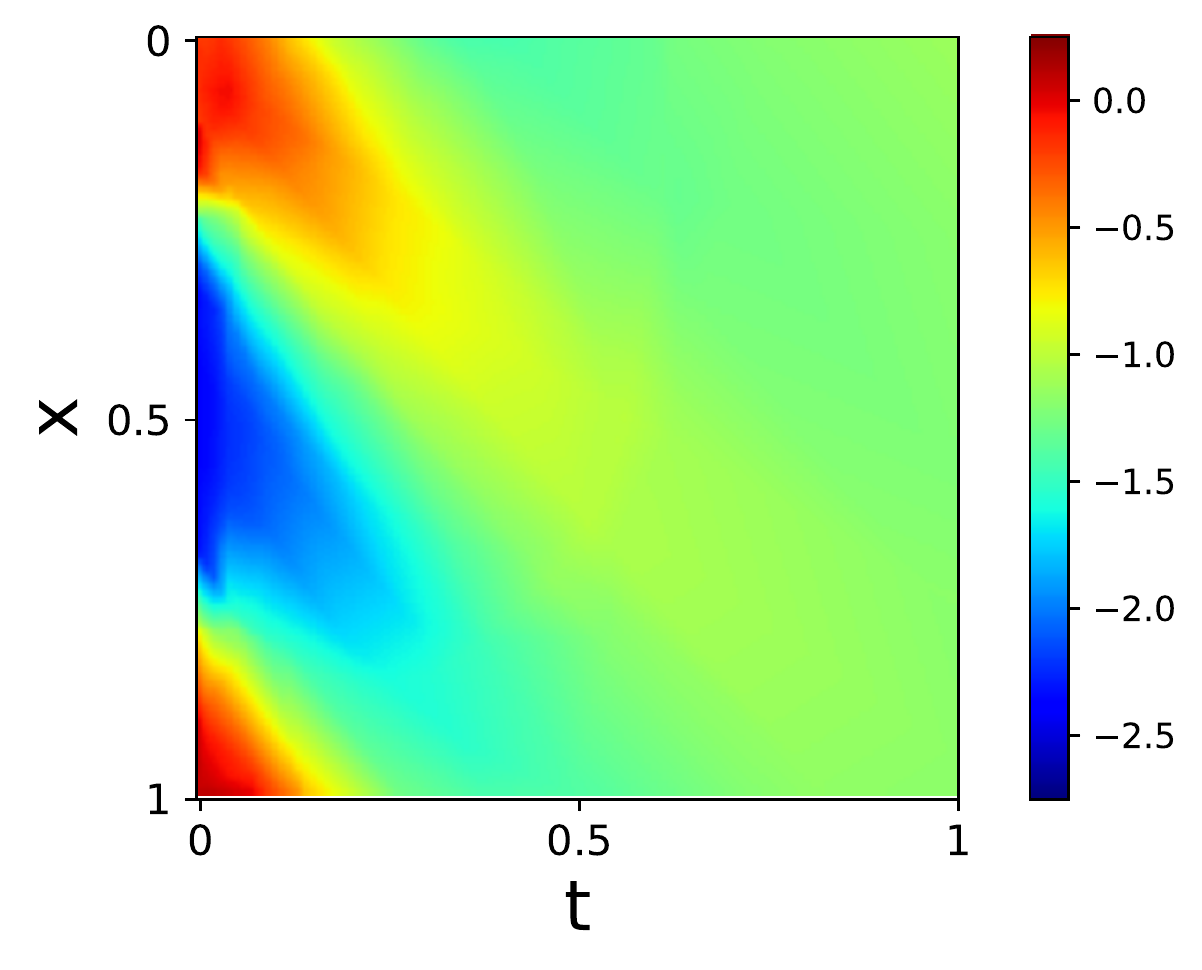}
\end{subfigure} &
\begin{subfigure}{0.245\textwidth}
\includegraphics[width = \textwidth]{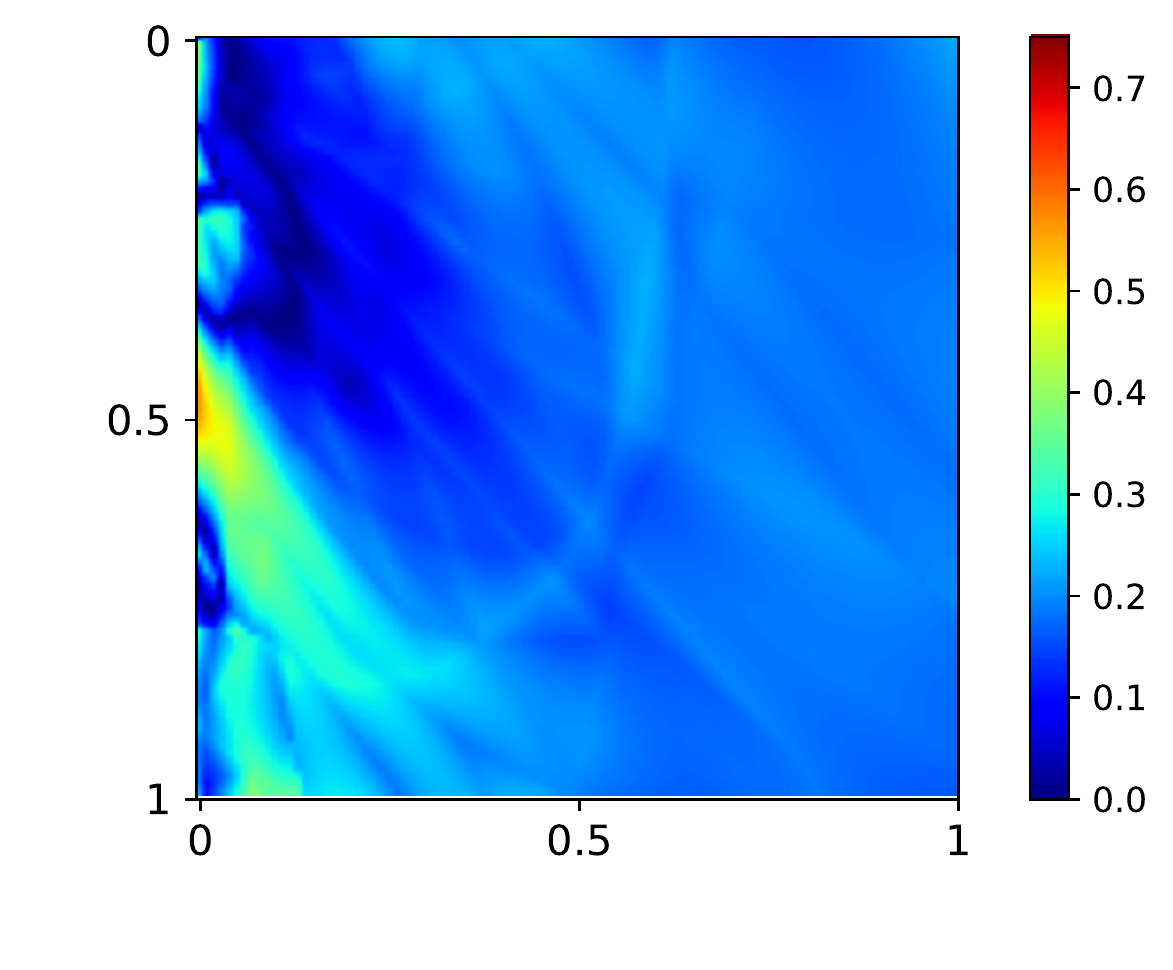}
\end{subfigure} &
\textbf{ - }\\[12pt]

VB-DeepONet &
\begin{subfigure}{0.245\textwidth}
\includegraphics[width = \textwidth]{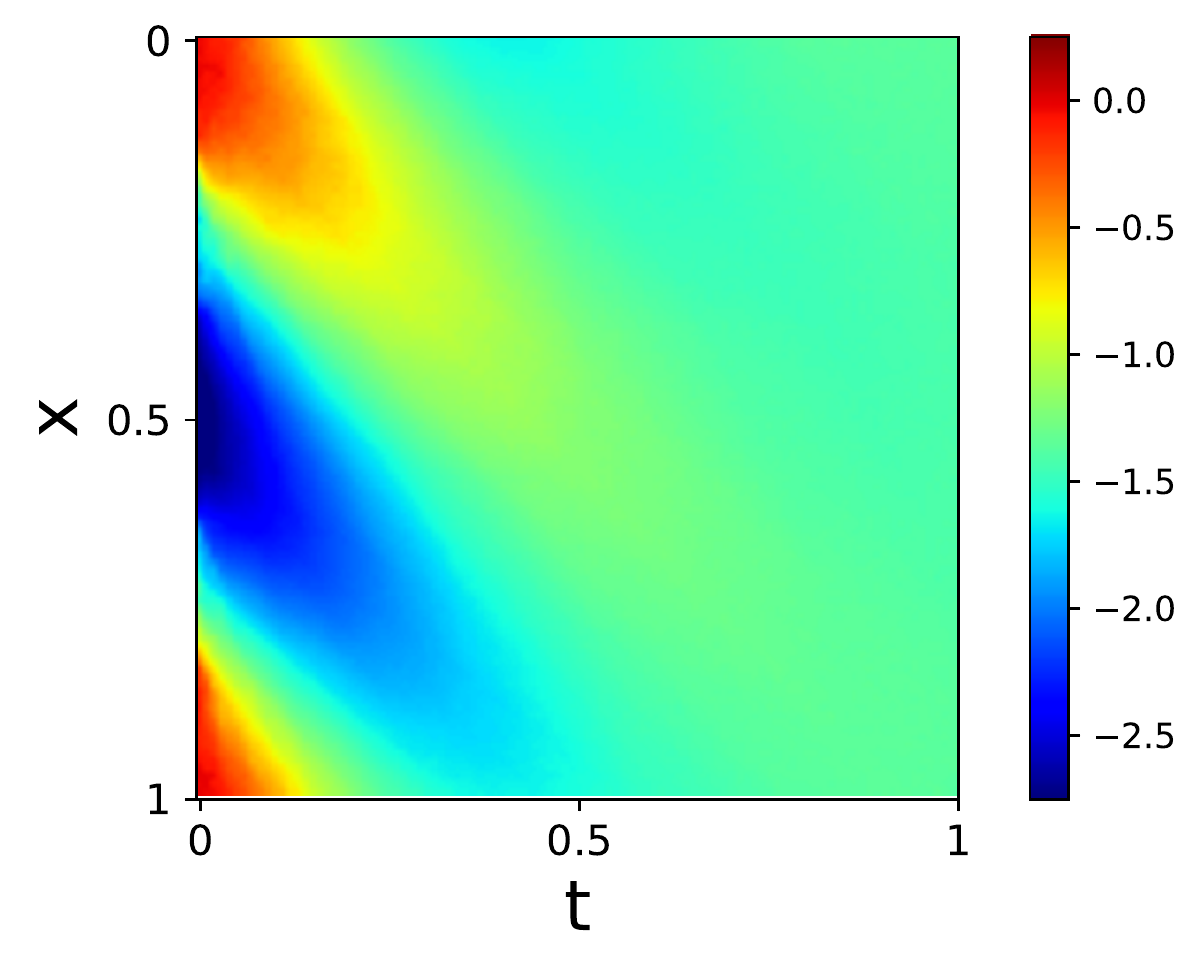}
\end{subfigure} &
\begin{subfigure}{0.245\textwidth}
\includegraphics[width = \textwidth]{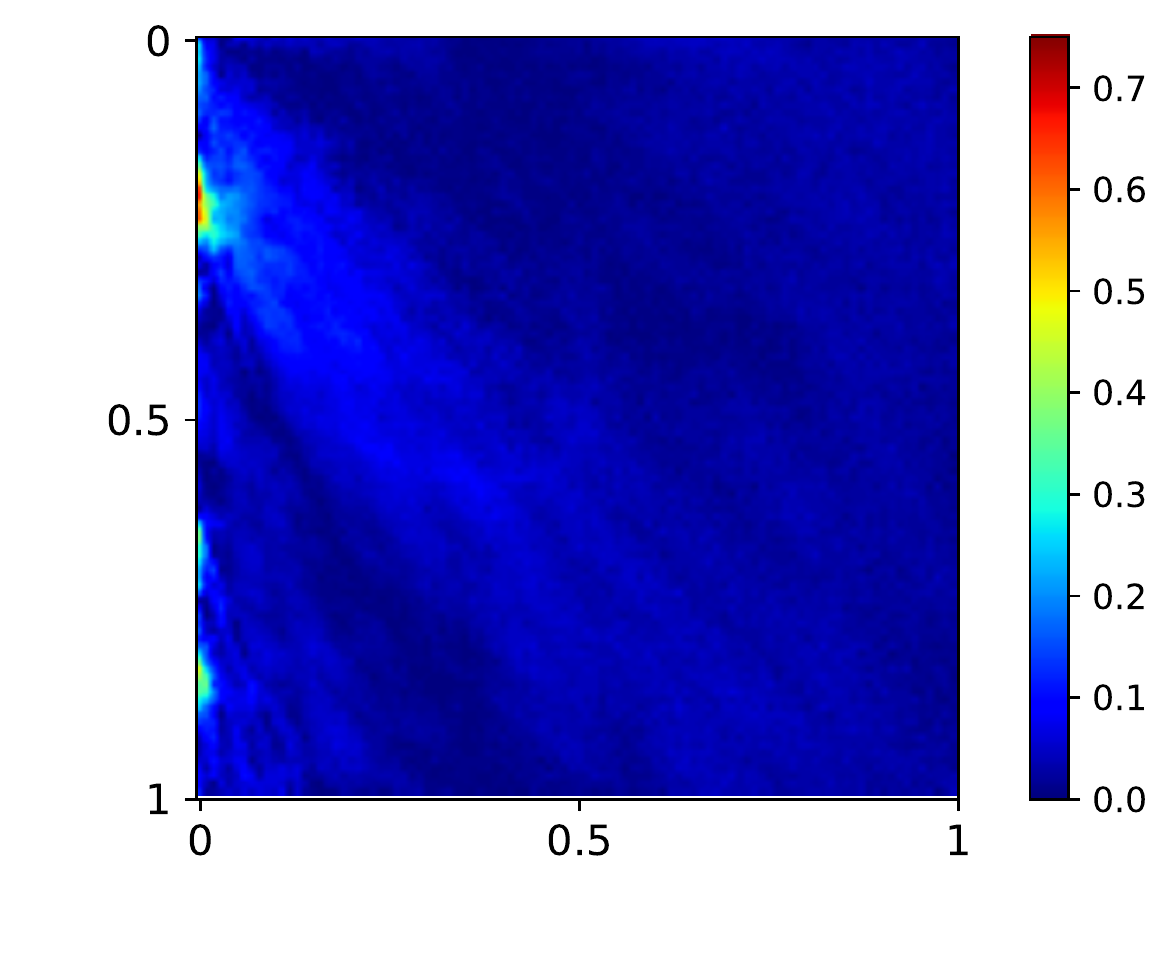}
\end{subfigure} &
\begin{subfigure}{0.245\textwidth}
\includegraphics[width = \textwidth]{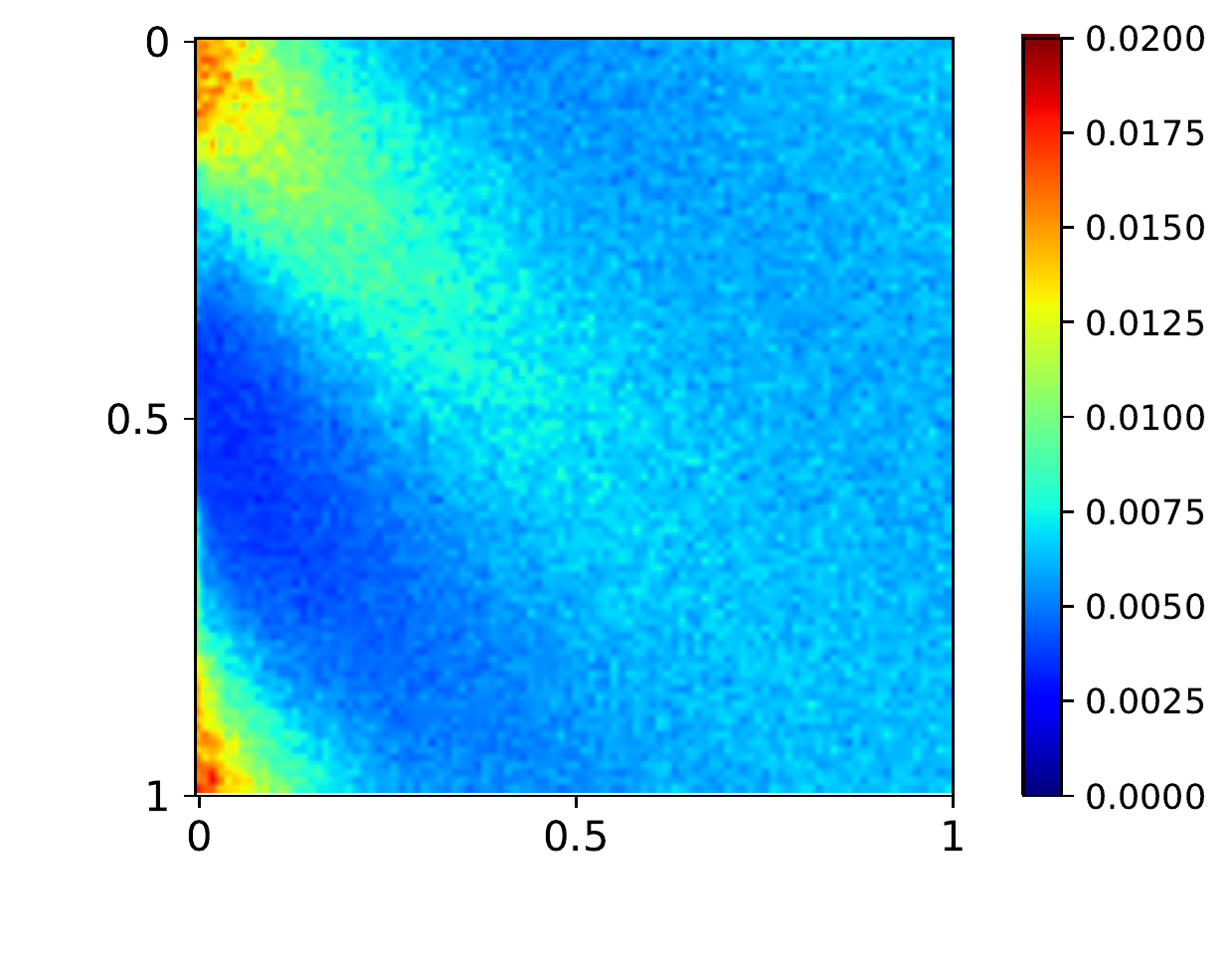}
\end{subfigure}\\
\hline
\end{tabular}
\vspace{0.25cm}
\caption{VB-DeepONet and D-DeepONet predictions compared against the ground truth for ADVD example.}
\label{table: CIV}
\end{table}
Fig. \ref{figure: CIV, CIs} shows the confidence interval limits along with the mean prediction for the VB-DeepONet and as can be observed, similar to previous example, the confidence interval is not arbitrarily large.
\begin{figure}[ht!]
\centering
\begin{subfigure}{0.32\textwidth}
\includegraphics[width = \textwidth]{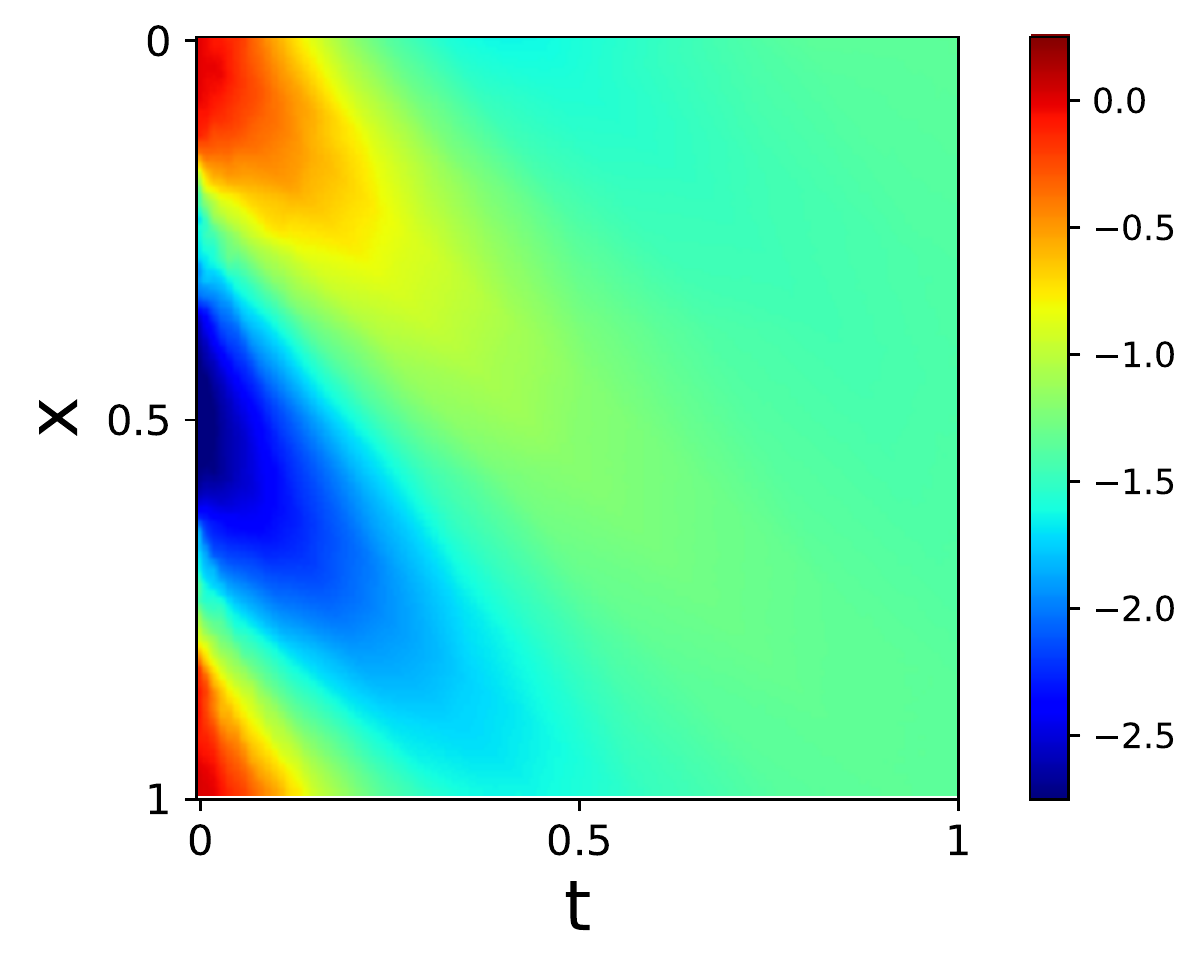}
\caption{\centering VB-DeepONet, 95\% CI, Upper Limit}
\end{subfigure}
\begin{subfigure}{0.32\textwidth}
\includegraphics[width = \textwidth]{ADVD_BDN_PRED.pdf}
\caption{VB-DeepONet, Mean Prediction}
\end{subfigure}
\begin{subfigure}{0.32\textwidth}
\includegraphics[width = \textwidth]{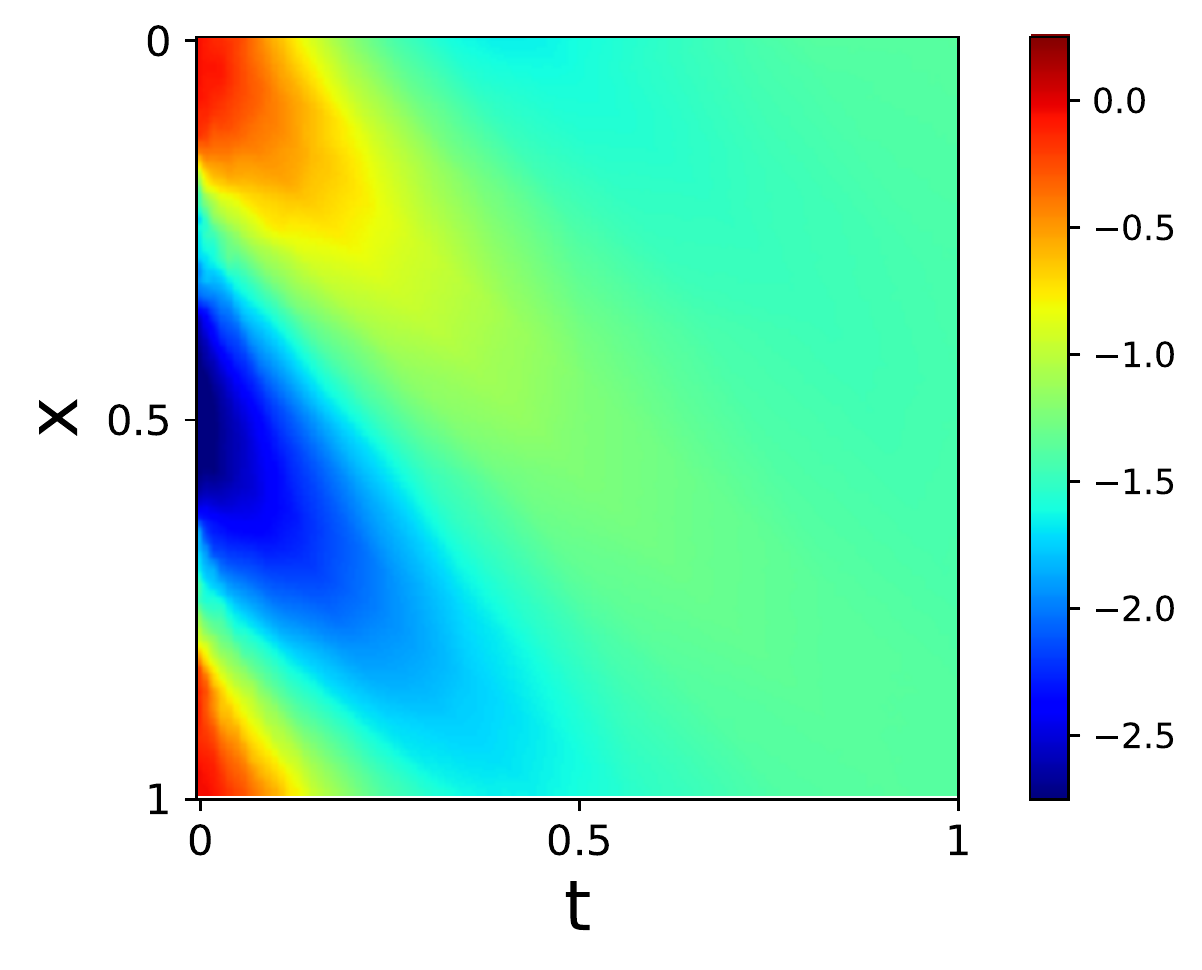}
\caption{\centering VB-DeepONet, 95\% CI, Lower Limit}
\end{subfigure}
\caption{VB-DeepONet results corresponding to single input realization for ADVD example.}
\label{figure: CIV, CIs}
\end{figure}

Fig. \ref{figure: CIV, PDFs} shows the PDFs plotted following the same procedure as in the previous example for solutions at $x = 0.24, t = 0.30$ and at $x = 0.82, t = 0.30$.
The mean PDF obtained using VB-DeepONet closely follow the ground truth and the confidence intervals impart extra flexibility while using the results in further applications.
As can be observed that VB-DeepONet far outshines the D-DeepONet for this example which maybe attributed to it being set in a Bayesian framework which naturally tackles issue such as overfitting.
\begin{figure}[ht!]
\centering
\begin{subfigure}{0.75\textwidth}
\includegraphics[width = \textwidth]{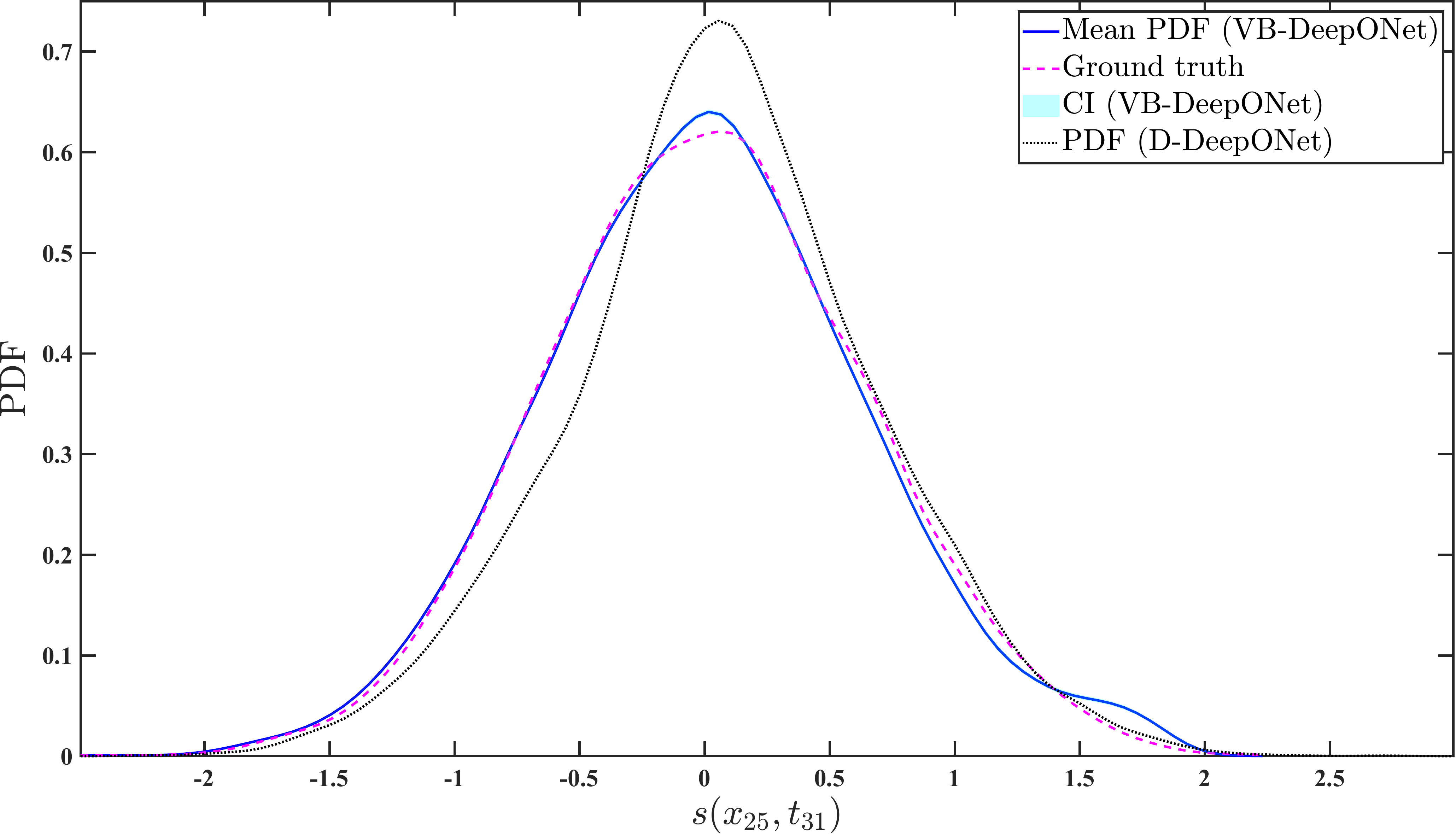}
\caption{PDF, $s(x_{24},t_{30})$}
\end{subfigure}
\begin{subfigure}{0.75\textwidth}
\includegraphics[width = \textwidth]{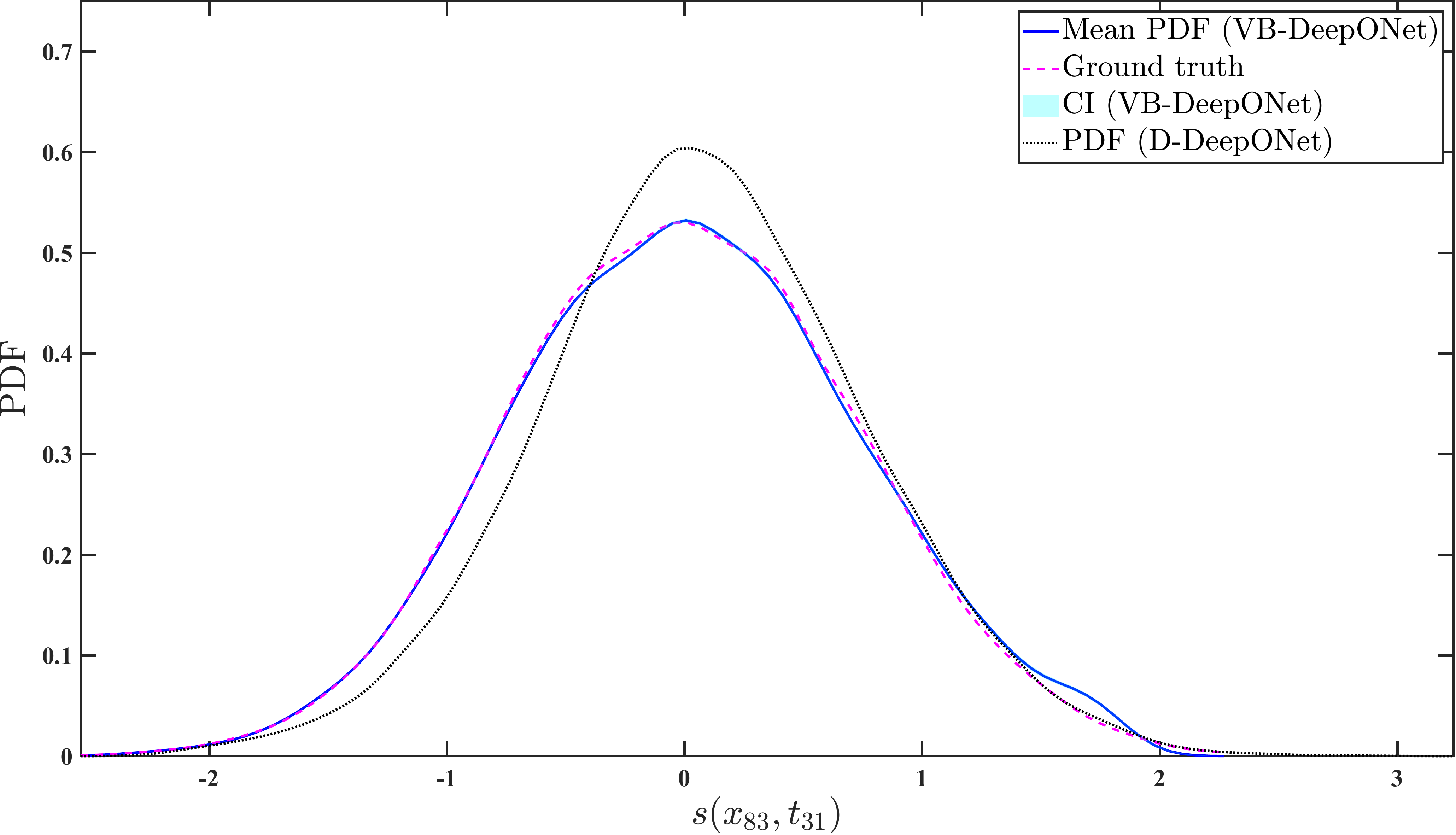}
\caption{PDF, $s(x_{82},t_{30})$}
\end{subfigure}
\caption{PDFs at different pairs of $x$ and $t$ for Advection-Diffusion example.}
\label{figure: CIV, PDFs}
\end{figure}

Before concluding this section, we summarize the results obtained using the proposed approach in a quantitative manner. To that end, we show the Normalized Mean Square Error (NMSE) for the three examples in Table \ref{table: nmse comparison}. The NMSE is calculated using the same 10000 realizations used in the prediction phase. 
For first, second, and the fourth examples, VB-DeepONet has shown better convergence than the D-DeepONet.
This may be because the VB-DeepONet is set in Bayesian framework and hence is less susceptible to overfitting and can therefore generalize better than D-DeepONet. 
For DR example, VB-DeepONet gives a NMSE of $0.610\%$, while the D-DeepONet performs slightly better with a NMSE of $0.458\%$.
However unequivocally, VB-DeepONet has the advantage of providing additional uncertainty quantification which is lacking in D-DeepONet.
\begin{table}[ht!]
\centering
\begin{tabular}{m{0.2\textwidth}m{0.125\textwidth}m{0.125\textwidth}m{0.125\textwidth}m{0.125\textwidth}}
\hline & Anti-derivative & Gravity Pendulum & Diffusion Reaction & Advection Diffusion \\ \hline
VB-DeepONet & 0.00009 & 0.00025 & 0.00610 & 0.00355 \\
DeepONet & 0.00016 & 0.00033 & 0.00458 & 0.00646\\
\hline
\end{tabular}
\vspace{0.25cm}
\caption{NMSE values obtained for VB-DeepONet predictions, compared against those obtained for D-DeepONet predictions.}
\label{table: nmse comparison}
\end{table}
\section{Conclusion}\label{section: conclusion}
The goal of this paper was to 
develop a Bayesian counterpart for the popular operator learning architecture, D-DeepONet.
The major difference in working in Bayesian framework is that the NN parameters are sampled from a probability distribution instead of being point estimates.
Working in Bayesian framework lends additional benefits like reducing the risk of overfitting which is something that deterministic NNs are highly susceptible to.

To reach the desired goal, one possible option was to use MCMC schemes to estimate the Bayesian posterior.
This however is computationally expensive.
Hence, in this paper, to develop the Bayesian DeepONet, variational inference has been used.
Variational inference is an approximate method which can significantly reduce the computational cost and can analyze deep NNs with large number of trainable parameters.
In variational inference, the Bayesian posterior to be estimated is replaced by a variational distribution and the loss function becomes the distance between the posterior and the variational distribution. The proposed framework is referred to as VB-DeepONet.
In the proposed VB-DeepONet framework, standard normal distribution is taken as the prior for the NN parameters and a normal distribution is taken as the variational distribution.

To illustrate the performance of the proposed framework, four different examples are shown, the first two of which follow ordinary DEs while the last two follow partial DEs.
The key observations are as follows:
\begin{itemize}
\item Overall, the proposed VB-DeepONet predicts more accurate results in three out of the four examples studies. This is probably because being Bayesian allows the framework to address the issue associated with overfitting.
\item VB-DeepONet being Bayesian yields confidence interval of the prediction as well. This is extremely important as it enables a decision maker to take more informed decision. 
\item On the hindsight, it is important to note that variational inference has the tendency to be overconfident. Therefore, the confidence interval obtained using variational inference is less as compared to a Monte Carlo based approach. This is evident in the PDF plot for the Advection Diffusion example where the true solution is outside the confidence interval. Nonetheless, for the other three examples, the ground truth is within the confidence interval.
\end{itemize}

Based on the overall results generated, it can be concluded that the proposed VB-DeepONet is a robust Bayesian operator learning framework.
Having said that, we note that even with variational inference, working in Bayesian framework is computationally more expensive than working with a deterministic framework. Also, a simple standard Gaussian prior is considered in this study. In future, a work can be carried out to address some of these issues.
\section*{Acknowledgment}
SG acknowledges the support received from the Ministry of Education in the form of Ph.D. scholarship.
SC acknowledges the financial support received from Science and Engineering Research Board (SERB) vi grant no. SRG/2021/000467 and from IIT Delhi in the form of seed grant.
\section*{Data availability}
On acceptance of the paper, the associated codes will be made available through github and the github link will be provided here.

\appendix
\section{VB-DeepONet architecture details}\label{appendix: NN architectures}
This section covers the details of the branch nets and trunk nets used in various examples.
The details of the VB-DeepONet used in the AD example are given in Table \ref{table: VBD E1}.
I(N) in Table \ref{table: VBD E1}-\ref{table: VBD E4} refers to the input layer with N nodes, D(N) refers to densely connected layer with N nodes, having random trainable parameters and ReLU denotes the \textit{Rectified Linear Unit} activation function.
\begin{table}[ht!]
\centering
\begin{tabular}{lll}
\hline
Net & Architecture \\
\hline
Branch & I(100) $\rightarrow$ D(30) $\rightarrow$ ReLU $\rightarrow$ D(30) $\rightarrow$ ReLU $\rightarrow$ D(30) $\rightarrow$ ReLU\\

Trunk & I(1) $\rightarrow$ D(30) $\rightarrow$ ReLU $\rightarrow$ D(30) $\rightarrow$ ReLU $\rightarrow$ D(30) $\rightarrow$ ReLU\\

Output & D(2)\\
\hline
\end{tabular}
\vspace{0.25cm}
\caption{VB-DeepONet details for the AD example.}
\label{table: VBD E1}
\end{table}
VB-DeepONet architecture details for GP example are given in Table \ref{table: VBD E2}.
\begin{table}[ht!]
\centering
\begin{tabular}{lll}
\hline
Net & Architecture \\
\hline
Branch & I(100) $\rightarrow$ D(25) $\rightarrow$ ReLU $\rightarrow$ D(25) $\rightarrow$ ReLU $\rightarrow$ D(25) $\rightarrow$ ReLU $\rightarrow$ D(25) $\rightarrow$ ReLU\\

Trunk & I(1) $\rightarrow$ D(25) $\rightarrow$ ReLU $\rightarrow$ D(25) $\rightarrow$ ReLU $\rightarrow$ D(25) $\rightarrow$ ReLU $\rightarrow$ D(25) $\rightarrow$ ReLU\\

Output & D(2)\\
\hline
\end{tabular}
\vspace{0.25cm}
\caption{VB-DeepONet details for the gravity pendulum example.}
\label{table: VBD E2}
\end{table}
VB-DeepONet architecture details for DR example are given in Table \ref{table: VBD E3}.
\begin{table}[ht!]
\centering
\begin{tabular}{lll}
\hline
Net & Architecture \\
\hline
Branch & I(100) $\rightarrow$ D(25) $\rightarrow$ ReLU $\rightarrow$ D(25) $\rightarrow$ ReLU $\rightarrow$ D(25) $\rightarrow$ ReLU $\rightarrow$ D(25) $\rightarrow$ ReLU\\

Trunk & I(2) $\rightarrow$ D(25) $\rightarrow$ ReLU $\rightarrow$ D(25) $\rightarrow$ ReLU $\rightarrow$ D(25) $\rightarrow$ ReLU $\rightarrow$ D(25) $\rightarrow$ ReLU\\

Output & D(2)\\
\hline
\end{tabular}
\vspace{0.25cm}
\caption{VB-DeepONet details for the DR example.}
\label{table: VBD E3}
\end{table}
VB-DeepONet architecture details for ADVD example are given in Table \ref{table: VBD E4}.
\begin{table}[ht!]
\centering
\begin{tabular}{lll}
\hline
Net & Architecture \\
\hline
Branch & I(100) $\rightarrow$ D(35) $\rightarrow$ ReLU $\rightarrow$ D(35) $\rightarrow$ ReLU $\rightarrow$ D(35) $\rightarrow$ ReLU\\

Trunk & I(2) $\rightarrow$ D(35) $\rightarrow$ ReLU $\rightarrow$ D(35) $\rightarrow$ ReLU $\rightarrow$ D(35) $\rightarrow$ ReLU\\

Output & D(2)\\
\hline
\end{tabular}
\vspace{0.25cm}
\caption{VB-DeepONet details for the ADVD example.}
\label{table: VBD E4}
\end{table}
\end{document}